\documentclass[sn-mathphys,Numbered]{sn-jnl}% Math and Physical Sciences Reference Style
\usepackage{graphicx}%
\usepackage{multirow}%
\usepackage{amsmath,amssymb,amsfonts}%
\usepackage{amsthm}%
\usepackage{mathrsfs}%
\usepackage[title]{appendix}%
\usepackage{xcolor}%
\usepackage{textcomp}%
\usepackage{manyfoot}%
\usepackage{booktabs}%
\usepackage{algorithm}%
\usepackage{algorithmicx}%
\usepackage{algpseudocode}%
\usepackage{listings}%
\usepackage[latin9]{inputenc}
\usepackage{xspace}
\usepackage{tabularx}
\usepackage{indentfirst}
\usepackage{amsmath}
\usepackage{amsthm}
\usepackage{microtype}
\usepackage{bm}
\usepackage{breakurl}
\usepackage[caption=false]{subfig}
\usepackage{times}
\usepackage[shortlabels]{enumitem}
\usepackage{tabularx}
\usepackage{url}
\usepackage{tikz}
\usepackage{pgflibraryarrows}
\usepackage{pgflibrarysnakes}
\usepackage{color} 
\usepackage{diagbox}
\usepackage{graphics}
\usepackage{multirow}
\usepackage{float} 
\usepackage{amsmath}
\usepackage{epsfig}
\usepackage{amssymb,amsfonts}
\usepackage{epstopdf}
\usepackage[T1]{fontenc}
\usepackage{changepage}
\usepackage{color,xcolor}

\usepackage{mlmath}
\newcommand{\bmv}{\bm{v}}

\newcommand{\TT}{\mathcal{T}}
\newcommand{\N}{\mathcal{N}}

\makeatletter
\newcommand*\bigcdot{\mathpalette\bigcdot@{.5}}
\newcommand*\bigcdot@[2]{\mathbin{\vcenter{\hbox{\scalebox{#2}{$\m@th#1\bullet$}}}}}
\makeatother

\newtheorem{thm}{Theorem}

\usepackage{lineno}

\theoremstyle{thmstyleone}%
\theoremstyle{thmstyletwo}%

\theoremstyle{thmstylethree}%
\usepackage{bbm}
\usepackage{url}
\newcommand{\black}[1]{{\color{black} #1}}

\begin{document}

\title[Article Title]{Overview frequency principle/spectral bias in deep learning\footnote{This paper is published in Communications on Applied Mathematics and Computation, and dedicated to the memory of Professor Zhong-Ci Shi.}}


\author*[1,2]{\fnm{Zhi-Qin} \spfx{John} \sur{Xu}}\email{xuzhiqin@sjtu.edu.cn}

\author[1,2]{\fnm{Yaoyu} \sur{Zhang}}\email{zhyy.sjtu@sjtu.edu.cn}

\author[2,1,3,4]{\fnm{Tao} \sur{Luo}}\email{luotao41@sjtu.edu.cn}

\affil[1]{\orgdiv{Institute of Natural Sciences,  MOE-LSC}, \orgname{Shanghai Jiao Tong University}, \orgaddress{\city{Shanghai}, \postcode{200240}, \country{China}}}
\affil[2]{\orgdiv{School of Mathematical Sciences}, \orgname{Shanghai Jiao Tong University}, \orgaddress{\city{Shanghai}, \postcode{200240}, \country{China}}}
\affil[3]{\orgdiv{CMA-Shanghai}, \orgname{Shanghai Jiao Tong University}, 
	\orgaddress{\city{Shanghai}, \postcode{200240}, \country{China}}}
\affil[4]{\orgname{Shanghai Artificial Intelligence Laboratory}, 
	\orgaddress{\city{Shanghai}, \postcode{200232}, \country{China}}}

% \affil[2]{\orgdiv{School of Mathematical Sciences, Institute of Natural Sciences, MOE-LSC and Qing Yuan Research Institute}, \orgname{Shanghai Jiao Tong University, Shanghai Center for Brain Science and Brain-Inspired Technology}, \\ 
%  \orgaddress{\city{Shanghai}, \postcode{200240}, \country{China}}}

% \affil[3]{\orgdiv{School of Mathematical Sciences, Institute of Natural Sciences, MOE-LSC and Qing Yuan Research Institute}, \orgname{Shanghai Jiao Tong University}, \\ \orgaddress{\city{Shanghai}, \postcode{200240}, \country{China}}}

% \author{%
%     % \name Tao\ Luo\footnotemark[1] \email luotao41@sjtu.edu.cn \\
%     \name Zhi-Qin John Xu\footnotemark[1] \email xuzhiqin@sjtu.edu.cn \\
%     \addr Institute of Natural Sciences, School of Mathematical Sciences, MOE-LSC and Qing Yuan Research Institute, \\Shanghai Jiao Tong University, Shanghai, 200240, China\\
%     \name Yaoyu Zhang \email zhyy.sjtu@sjtu.edu.cn \\
%     \addr School of Mathematical Sciences, Institute of Natural Sciences, MOE-LSC and Qing Yuan Research Institute,\\ Shanghai Jiao Tong University, Shanghai Center for Brain Science and Brain-Inspired Technology, Shanghai, 200240, China \\
%     \name Tao Luo \email luotao41@sjtu.edu.cn \\
%     \addr School of Mathematical Sciences, Institute of Natural Sciences, MOE-LSC and Qing Yuan Research Institute, Shanghai Jiao Tong University, Shanghai, 200240, China }


\abstract{Understanding deep learning is increasingly emergent as it penetrates more and more into industry and science. In recent years, a research line from Fourier analysis sheds lights on this magical ``black box'' by showing a Frequency Principle (F-Principle or spectral bias) of the training behavior of deep neural networks (DNNs) --- DNNs often fit functions from low to high frequencies during the training. The F-Principle is first demonstrated by one-dimensional synthetic data followed by the verification in high-dimensional real datasets. A series of works subsequently enhance the validity of the F-Principle. This low-frequency implicit bias reveals the strength of neural network in learning low-frequency functions as well as its deficiency in learning high-frequency functions. Such understanding inspires the design of DNN-based algorithms in practical problems, explains experimental phenomena emerging in various scenarios, and further advances the study of deep learning from the frequency perspective. Although incomplete, we provide an overview of F-Principle and propose some open problems for future research.}




\maketitle

\section{Introduction}
\subsection{Motivation}
In practice, deep learning, often realized by deep neural networks (DNNs), has achieved tremendous success in many applications, such as computer vision, speech recognition, speech translation, and natural language processing, etc. It also has become an indispensable method for solving a variety of scientific problems. On the other hand, DNN sometimes fails and causes critical issues in applications. In theory, DNN remains a black box for decades. Many researchers make the analogy between the practical study of DNN and the alchemy. Due to the booming application of DNNs, it has become an important and urgent mission to establish a better theoretical understanding of DNNs.

In recent years, theoretical study of DNNs has flourished. Yet, we still need clear demonstration of how these theoretical results provides key insight and guidance to practical study of DNNs. An insightful theory usually provide guidance for practice from two aspects\---capability and limitation. For example, the conservation of mass in chemistry informs the fundamental limitation of chemical reactions that they cannot turn one element into another, e.g., turning bronze into gold. On the other hand, they may combine elementary substances into their compounds. These understandings are extremely valuable, with which, the study of alchemy transforms into modern chemistry. In this work, we overview the discovery and studies about the frequency principle of deep learning \citep{xu_training_2018,xu2019frequency,rahaman2018spectral,zhang2021linear}, by which we obtain a basic understanding of the capability and limitation of deep learning, i.e., the difficulty in learning and achieving good generalization for high frequency data as well as the easiness and intrinsic preference for low frequency data. Based on this guideline of frequency principle, many algorithms have been developed to either employ this low frequency bias of DNN to well fit smooth data or design special tricks/architectures to alleviate the difficulty of DNN in fitting data known to be highly oscillatory \citep{liu2020multi,jagtap2019adaptive,cai2019phasednn,tancik2020fourier}. Hopefully, by the development of frequency principle and theories from other perspectives, the practical study of deep learning would become a real science in the near future.

The discovery of frequency principle is made to confront the following open puzzle central for DNN theories: why over-parameterized DNNs generalize well in many problems, such as natural image classification.
% why stochastic gradient descent can train a DNN with huge number of parameters, in which the loss landscape is extremely complicate; What is the limitation of DNNs; etc.
% Since 1990s, the universal approximation theorem \citep{cybenko1989approximation} has shown that over-parameterized DNNs have sufficient capacity in approximating continuous functions. However, one cares more about the prediction ability, i.e., generalization error, than the approximation ability in practice. In addition, the solution constructed in the approximation proof is highly impossible to be found by a training algorithm. 
In 1995, Vladimir Vapnik and Larry Jackel made a bet, witnessed by Yann LeCun, that is, Larry claimed that by 2000 we would have had a theoretical understanding of why big neural nets work well (in the form of a bound similar to what we have for SVMs). Also in 1995, Leo Breiman published a reflection after refereeing papers for NIPS \citep{breiman1995reflections}, where he raised many important questions regarding DNNs, including ``why don't heavily parameterized neural networks overfit the data''. 
% What is the effective number of parameters? Why doesn't backpropagation head for a poor local minima? When should one stop the backpropagation and use the current parameters? Today, all these issues proposed in 1995 are still largely unanswered. 
In 2016, an empirical study \citep{zhang2016understanding} raises much attention again to this over-parameterization puzzle with systematic demonstration on modern DNN architectures and datasets.
% , i.e., Why don't heavily parameterized neural networks overfit the data. 
This over-parameterization puzzle contradicts the conventional generalization theory and traditional wisdom in modeling, which suggests that a model of too many parameters easily overfits the data. This is exemplified by von Neumann's famous quote ``with four parameters I can fit an elephant" \citep{dyson_meeting_2004}. Establishing a good theoretical understanding of this over-parameterization puzzle has since become more and more crucial as modern DNN architecture incorporates increasingly more parameters, e.g., $\sim 10^8$ for VGG19 \citep{simonyan2014very}, $\sim 10^{11}$ for GPT-3 \citep{brown2020language}, which indeed achieves huge success in practice.

To address this puzzle, a notable line of works, starting from the conventional complexity-based generalization theory, attempt to propose novel norm-based complexity measures suitable for DNNs.
% To study the generalization in deep learning, a series of works propose norm-based complexity measures, 
However, a recent empirical study shows that many norm-based complexity measures not only perform poorly, but negatively correlate with generalization, specifically, when the optimization procedure injects some stochasticity \citep{jiang2019fantastic}. 
% Generalization error bound characterized by complexity measures, such as Rademacher complexity \citep{bartlett2002rademacher}, consider the worst case. 
% For DNNs, the fitting depends on many factors, such as initialization, training algorithm, network structure, etc. It is unclear how these factors affect the fitting, thus, it is difficult to find a tight estimation of the worst case or a complexity measure. 
Another line of works start from a variety of ideal models of DNNs, e.g., deep linear network \citep{saxe_exact_2013,saxe_information_2019,lampinen_analytic_2019}, committee machine \citep{engel_statistical_2001,aubin_committee_2018}, spin glass model \citep{choromanska_loss_2015}, mean-field model \citep{mei2018mean,rotskoff_parameters_2018,chizat_global_2018,sirignano_mean_2020}, neural tangent kernel \citep{jacot2018neural,lee_wide_2019}. These works emphasize fully rigorous mathematical proofs and have difficulties in providing a satisfactory explanation for general DNNs \citep{zdeborova2020understanding}.  

% DNNs are very different from conventional optimization methods. To solve a conventional optimization problem with multiple solutions, one often explicitly adds a penalty term. For example, to obtain a sparse solution, one can impose a $L_1$ regularization. To train an over-parameterizedDNN, which also has multiple solution to achieve a satisfactory training error, one needs to select a loss function (e.g., mean squared error), a proper initialization for weights (e.g., LeCun initialization), and an optimization algorithm (e.g., gradient descent or stochastic gradient descent). These factors add up to a trajectory that leads to an unique solution. Compared with conventional optimization problems, one does not require an explicit regularization term and the training trajectory on the loss landscape matters. \zyy{If all these factors in training DNNs can be equivalent to a combination of the loss function with a regularization term, then, the regularization term would render the characteristics of the DNN methods, that is, understanding what kind of problems DNNs can generalize well. Since this regularization term is not explicitly imposed, it is also called implicit regularization or implicit bias. A key to understand the implicit regularization of DNNs is the training trajectory. } 

\begin{figure}
	\centering
	\subfloat[]{
		\includegraphics[width=0.33\textwidth]{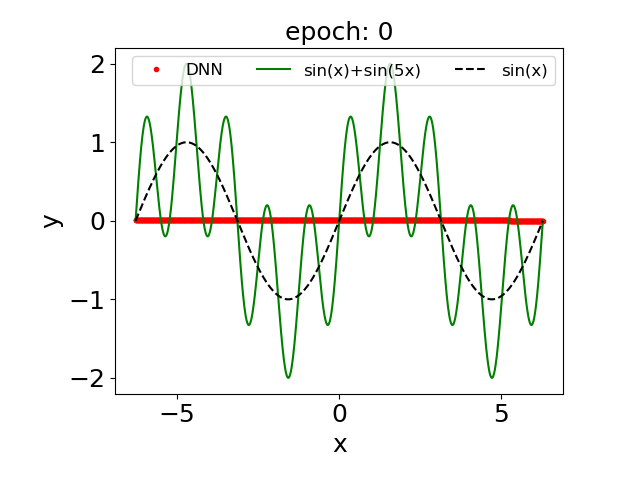}
	}
	\subfloat[]{
		\includegraphics[width=0.33\textwidth]{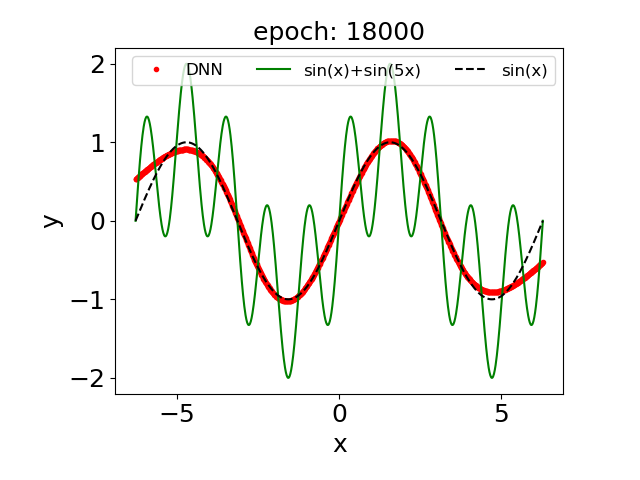}
	}
	\subfloat[]{
		\includegraphics[width=0.33\textwidth]{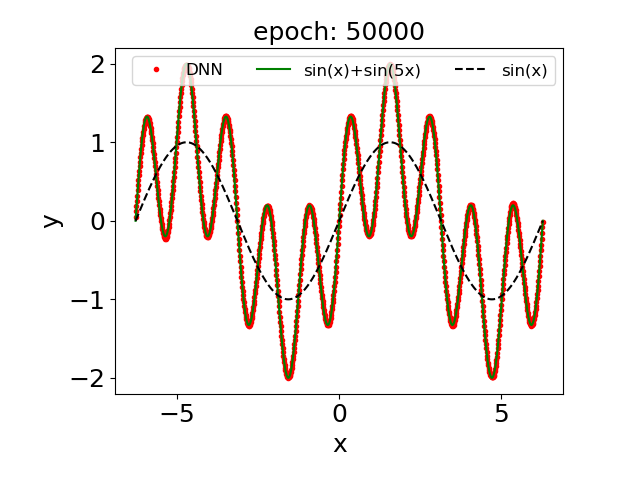}
	}
	\centering
	\caption{\label{fig:1dfp}Illustration of the training process of a DNN. Training data are sampled from target function $\sin(x)+\sin(5x)$. Red, green and black curves indicates DNN output, $\sin(x)$, and $\sin(x)+\sin(5x)$ respectively.}
\end{figure}

The frequency principle overviewed in this paper takes a phenomenological approach to  deep learning theory, to understand complex systems, black boxes at first glance, in science and especially in physics. 
% The frequency principle/spectral bias overviewed in this paper takes another approach to study the training trajectory of DNNs, i.e., a combination of phenomenological study and theoretical study.  
Taking this approach, the first difficulty we encounter is the extreme complexity of deep learning experiments in practice. For example, the MNIST dataset is a well-known simple (if not too simple) benchmark for testing a DNN. However, the learned DNN is already a very high dimensional (784-dimensional) mapping, which is impossible to be visualized and analyzed exactly.
% DNNs prevail in solving high-dimensional problems, which are notoriously difficult for theoretical analysis and visualization. 
% The dataset MNIST is a well-known and very simple benchmark example. However, the training of a DNN on MNIST is to learn a function mapping from 784-dimensional space to 10-dimensional space. The analysis of such high-dimensional functions is often extremely difficult. 
In face of such difficulty, an important step we take is to carefully design synthetic problems simple enough for thorough analysis and visualization of the DNN learning process, but complicated enough for reproducing interested phenomena. 
We train DNNs to fit a function with one-dimensional input and one-dimensional output like $\sin(x)+\sin(5x)$ shown in Fig. \ref{fig:1dfp}. Luckily, a clear phenomenon emerges from the thorough visualization of the DNN training process that the DNN first captures a coarse and relatively ``flat'' landscape of the target function, followed by more and more oscillatory details. It seems that the training of a DNN gives priority to the flat functions, which should generalize better for flat target function by intuition, over the oscillatory functions. By the phenomenological approach, we next quantify this phenomenon by the Fourier analysis, which is a natural tool to quantify flatness and oscillation. As shown later,  by transforming the DNN output function into the frequency domain, the differences in convergence rate between flat and oscillatory components become apparent. 
% , which is a natural tool to quantify flatness and oscillation in detail in mathematics and signal processing.
% The oscillation and flatness discussed here is also related to the generalization issue, that is, a flat function often generalizes better. 
% Therefore, this synthetic problem motives us to study the training trajectory from frequency domain, which is natural to describe flatness and oscillation.
We conclude this phenomenon of implicit low-frequency bias by the frequency principle/spectral bias \citep{xu_training_2018,xu2019frequency,rahaman2018spectral,zhang2021linear}, i.e., DNNs often fit target functions from low to high frequencies during the training, followed by extended experimental studies for real datasets and a series of theoretical studies detailed in the main text.  

In the end, as a reflection, we note that the specialness of the discovery of frequency principle lies in our faith and insistence on performing systematic DNN experiments on simple $1$-d synthetic problems, which is simple for observation and analysis but not clearly understood. From the perspective of phenomenological study, such simple cases serve as an excellent starting point, however, they are rarely considered in the experimental studies of DNNs. Some researchers even deems MNIST experiments as too simple for an empirical study without realizing that even phenomenon regarding the training of DNN on $1$-d problems is not well studied. In addition, since Fourier analysis is not naturally considered for high dimensional problems due to the curse of dimensionality, it is difficult even to think about Fourier transform for DNNs on real datasets as done in Sec. \ref{sec:empirical} without making a direct observation of DNN learning from flat to oscillatory in 1-d problems. Therefore, by overviewing the discovery and studies of frequency principle, we advocate for the phenomenological approach to the deep learning theory, by which systematic experimental study on simple problems should be encouraged and serve as a key step for developing the theory of deep learning.

%This paper reviews an understanding of DNNs from frequency perspective. An initial motivation of utilizing the Fourier analysis is to study the generalization and training speed of DNNs, which are central issues in deep learning. 
%The study of generalization in deep learning attracts much attention in recent years due to its contradiction to the traditional wisdom \citep{breiman1995reflections,zhang2016understanding}. Traditional wisdom suggests that a model of too many parameters easily overfit the data. However, over-parameterized DNNs often generalize well in practice \citep{zhang2016understanding}.  

\subsection{Frequency principle}
To visualize or characterize the training process in frequency domain requires a Fourier transform of the training data. However, the Fourier transform of high-dimensional data suffers from the curse of dimensionality and the visualization of high-dimensional data is difficult. Alternatively, one can study the problem of one-dimensional synthetic data.  \black{A series of experiments on synthetic low-dimensional data show that the DNN training follows a frequency principle (F-Principle) \citep{xu_training_2018,xu2019frequency,zhang2021linear}, that is,}

\begin{changemargin}{0.5cm}{0.5cm}\emph{DNNs often fit target
functions from low to high frequencies during the training.} 
\end{changemargin}
This implicit frequency bias is also called spectral bias \citep{rahaman2018spectral} and can be robustly observed no matter how overparameterized DNNs are. More experiments on real datasets are designed to confirm this observation \citep{xu2019frequency}. It is worthy to note that the frequency used here is a response frequency characterizing how the output is affected by the input. This frequency is easy to be confused in imaging classification problems. For example, in MNIST dataset, the frequency domain is also 784-dimensional but not 2-dimensional, i.e., the frequency of the classification function but not the image frequency w.r.t. 2-dimensional space. 

\citet{xu2019frequency} proposed a key mechanism of the F-Principle that the regularity of the activation function  converts into the decay rate of a loss function in the frequency domain. Theoretical studies subsequently show that the F-Principle holds in general setting with \black{continuous} samples \citep{luo2019theory} and in the regime of wide DNNs (Neural Tangent Kernel (NTK) regime \citep{jacot2018neural}) with finite samples \citep{zhang2019explicitizing,zhang2021linear,luo2020theory} or samples distributed uniformly on sphere \citep{cao2019towards,yang2019fine,basri2019convergence,bordelon2020spectrum}. \black{\citet{e2019machine} study the neural network from a continuous viewpoint, where neurons are treated as a discrete version of a continuous distribution of weights, and samples as a discrete version of another continuous distribution of data. The evolution of the network output during the training follows an integral equation, which would naturally lead  to the training that follows F-Principle}. In addition to characterizing the training speed of DNNs, the F-Principle also implicates that DNNs prefer low-frequency function and generalize well for low-frequency functions \citep{xu2019frequency,zhang2019explicitizing,zhang2021linear,luo2020theory}. 
 
The F-Principle further inspires the design of DNNs to fast learn a function with high frequency, such as in scientific computing and image or point cloud fitting problems \citep{liu2020multi,jagtap2019adaptive,cai2019phasednn,tancik2020fourier}. In addition, the F-Principle provides a mechanism to understand many phenomena in applications and inspires a series of studies on deep learning from frequency perspective. The study of deep learning is a highly inter-disciplinary problem. As an example, the Fourier analysis, an approach of signal processing, is an useful tool to better understand deep learning \citep{giryes2020how}. A comprehensive understanding of deep learning remains an exciting research subject calling for more fusion of existing approaches and new methods.

\section{Empirical study of F-Principle\label{sec:empirical}}
Before the discovery of the F-Principle, some works have suggested the learning of the DNNs may follow a order from simple to complex \citep{arpit2017closer}. However, previous empirical studies focus on the real dataset, which is high-dimensional. Thus, it is difficult to find a suitable quantity to characterize such intuition. In this section, we review the empirical study of the F-Principle, which first presents a clear picture from the one-dimensional data  and then carefully designs experiments to verify the F-Principle in high-dimensional data \citep{xu_training_2018,xu2019frequency,rahaman2018spectral}. 

\subsection{Frequency principle in low-dimensional problems}
To clearly illustrate the phenomenon \black{that follows F-Principle}, one can use $1$-d synthetic data to show the relative error of different frequencies during the training of DNN. The following shows an example from \citet{xu2019frequency}.

Training samples are drawn from a $1$-d target function $f(x)=\sin(x)+\sin(3x)+\sin(5x)$ with three important frequency components and even space in $[-3.14,3.14]$, i.e., $\{x_{i},f(x_{i})\}_{i=0}^{n-1}$. The discrete Fourier transform (DFT) of $f(x)$ or the DNN output (denoted by $h(x)$) is computed by $\hat{f}_{k}=\frac{1}{n}\sum_{i=0}^{n-1}f(x_{i})\E^{-\I2\pi ik/n}$, where $k$ is the frequency. As shown in Fig.~\ref{fig:onelayer}(a), the target function has three important frequencies as designed (black dots at the inset in Fig.~\ref{fig:onelayer}(a)). \black{We use a network with four hidden layers consisting of 200, 200, 200, 100 neurons, respectively. Both weights and bias are initialized from a uniform distribution on $[-\sqrt{1/m_{in}},\sqrt{1/m_{in}}]$, where $m_{in}$ is the number of input neurons. To examine the convergence behavior of different frequency components during the training with MSE and gradient descent}, we compute the relative difference between the DNN output and the target function for the three important frequencies at each recording step, that is, $\Delta_{F}(k)=|\hat{h}_{k}-\hat{f}_{k}|/|\hat{f}_{k}|$, where $|\cdot|$ denotes the norm of a complex number. As shown in Fig.~\ref{fig:onelayer}(b), the DNN converges the first frequency peak very fast, while converging the second frequency peak much more slowly, followed by the third frequency peak.

A series of experiments are performed with relative cheap cost on synthetic data to verify the validity of the F-Principle and eliminate some misleading factors. For example,  the stochasticity and the learning rate are not important to reproduce \black{the spectral bias phenomenon that follows F-Principle}. If one only focuses on high-dimensional data, such as the simple MNIST, it would require a much more expensive cost of computation and computer memory to examine the impact of the stochasticity and the learning rate. The study of synthetic data shows a clear guidance to examine the F-Principle in the high-dimensional data. In addition, since frequency is a quantity which theoretical study is relatively easy to access, the F-Principle provides a theoretical direction for further study.

\begin{center}
\begin{figure}
\begin{centering}
\includegraphics[width=\textwidth]{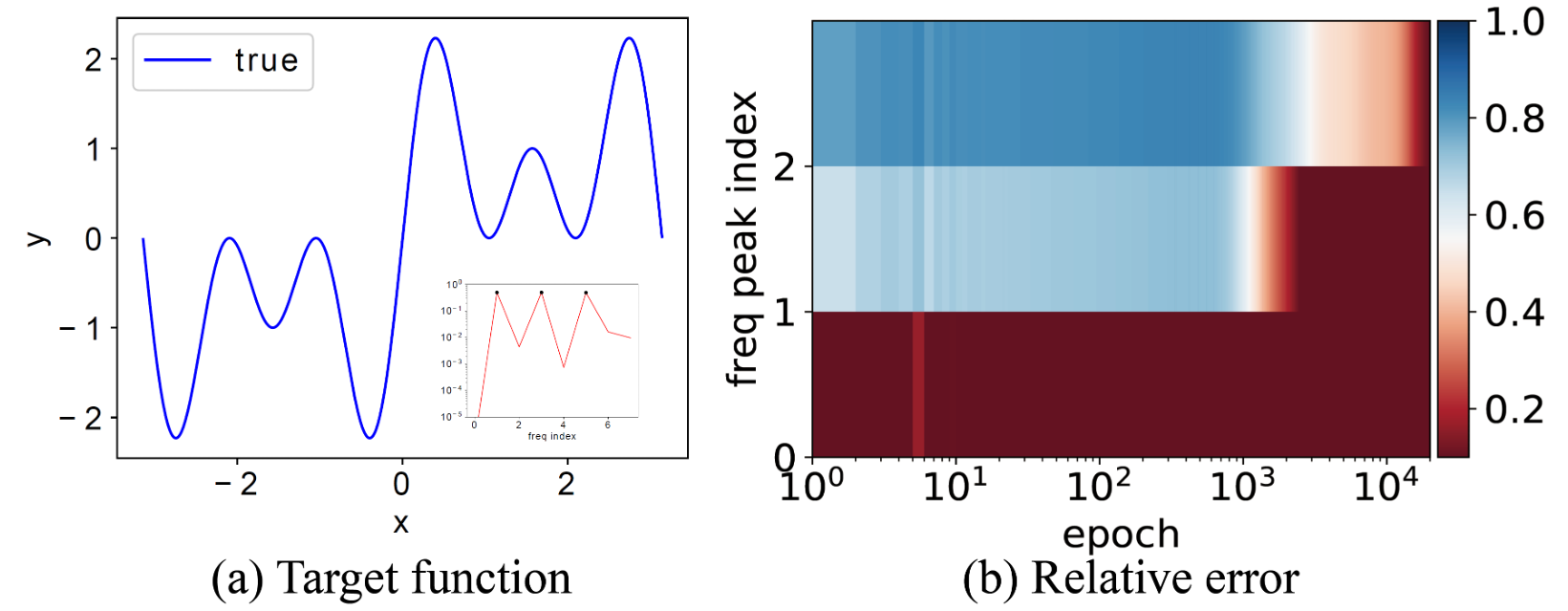} 
\par\end{centering}
\caption{1d input. (a) $f(x)$. Inset : $|\hat{f}(k)|$. (b) $\Delta_{F}(k)$
of three important frequencies (indicated by black dots in the inset
of (a)) against different training epochs. Reprinted from \citet{xu2019frequency}. \label{fig:onelayer} }
\end{figure}
\par\end{center}
% \subsection{Two-dimensional experiments}
An image can be regarded as a mapping from two-dimensional space coordinate to pixel intensity. \black{Learning this problem with a mean squared loss is a two-dimensional regression problem.} The experiment in Fig.  \ref{fig:2d} uses a fully-connected DNN to fit the camera-man image in Fig.  \ref{fig:2d}(a). The DNN learns from a coarse-grained image to produce one with more details as the training proceeds, shown in Fig. \ref{fig:2d}(b-d). Obviously, this is also an order from low- to high-frequencies, which is similar to how biological brain remembers an image. \black{In \citet{xu2018understanding}, the F-Principle is examined through the quantitative characterization of the convergence of each frequency in a cross section of a two-dimensional regression problem.}

This 2-d example also shows that utilizing DNN to restore an image may take advantage of low-frequency preference, such as inpainting tasks, but also should be cautionary about its insufficiency in learning high-frequency structures. To overcome this insufficiency, some algorithms are developed  \citep{chen2020ssd,jiang2020focal,xi2020drl,tancik2020fourier}, which will be reviewed more in Section \ref{sec:inspirealg}.

\begin{center}
\begin{figure}[h]
\begin{centering}
\subfloat[True]{\begin{centering}
\includegraphics[width=0.24\textwidth]{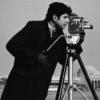} 
\par\end{centering}
}\subfloat[step 80]{\begin{centering}
\includegraphics[width=0.24\textwidth]{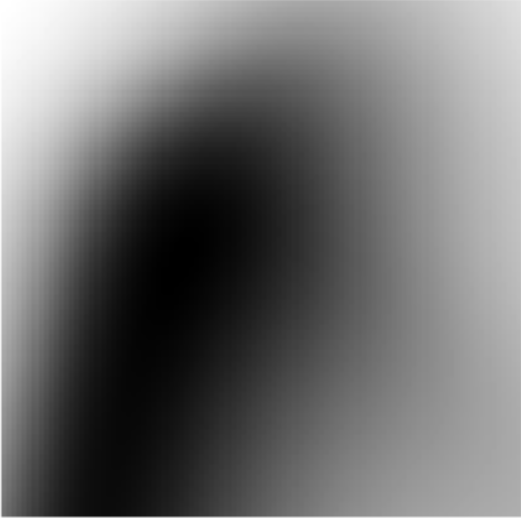} 
\par\end{centering}
}\subfloat[step 2000]{\begin{centering}
\includegraphics[width=0.24\textwidth]{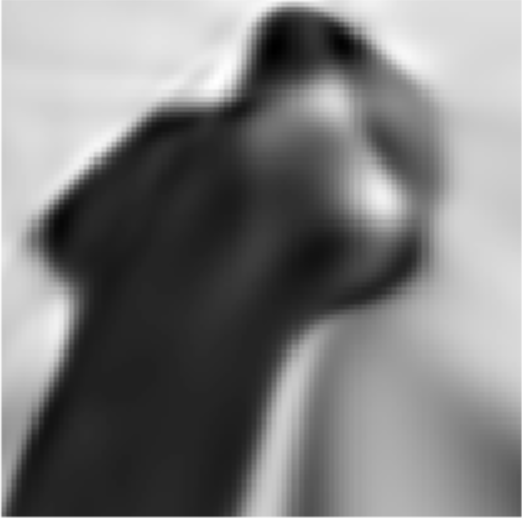} 
\par\end{centering}
}\subfloat[step 58000]{\begin{centering}
\includegraphics[width=0.24\textwidth]{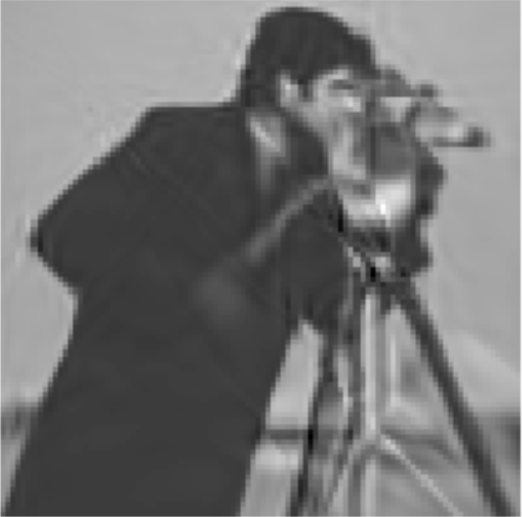} 
\par\end{centering}
}
\par\end{centering}
\caption{F-Principle in 2-d datasets.  Reprinted from \citet{xu2019frequency}. \label{fig:2d} }
\end{figure}
\par\end{center}

\subsection{Frequency principle in high-dimensional problems}
To study the F-Principle in high-dimensional data, two obstacles should be overcome first: what is the frequency in high dimension and how to separate different frequencies.

The concept of ``frequency'' often causes confusion in image classification problems. The image (or input) frequency (NOT used in \black{studying} F-Principle of classification problems) is the frequency of $2$-d
function $I:\mathbb{R}^{2}\to\mathbb{R}$ representing the intensity
of an image over pixels at different locations. This frequency corresponds
to the rate of change of intensity \emph{across neighbouring pixels}.
For example, an image of constant intensity possesses only the zero
frequency, i.e., the lowest frequency, while a sharp edge contributes
to high frequencies of the image. 

The frequency used in \black{studying} F-Principle of classification problems is also called 
\textbf{response frequency} of a general
Input-Output mapping $f$. For example, consider a simplified classification
problem of partial MNIST data using only the data with label $0$
and $1$, $f(x_{1},x_{2},\cdots,x_{784}):\mathbb{R}^{784}\to\{0,1\}$
mapping $784$-d space of pixel values to $1$-d space, where $x_{j}$
is the intensity of the $j$-th pixel. Denote the mapping's Fourier
transform as $\hat{f}(k_{1},k_{2},\cdots,k_{784})$. The frequency
in the coordinate $k_{j}$ measures the rate of change of $f(x_{1},x_{2},\cdots,x_{784})$
\emph{with respect to $x_{j}$, i.e., the intensity of the $j$-th
pixel}. If $f$ possesses significant high frequencies for large $k_{j}$,
then a small change of $x_{j}$  in the image
might induce a large change of the output (e.g., adversarial example). For a real data, the response frequency is rigorously defined via the standard nonuniform discrete Fourier transform (NUDFT).

The difficulty of separating different frequencies is that the computation of Fourier transform of high-dimensional data suffers from the curse of dimensionality. For example, if one evaluates two points in each dimension of frequency space, then, the evaluation of the Fourier transform of a $d$-dimensional function is on $2^d$ points, an impossible large number even for MNIST data with $d=784$. Two approaches are proposed in \citet{xu2019frequency}. 

\subsubsection{Projection method \label{sec:projectmethod}}

One approach is to study the frequency in one dimensional frequency space. 
For a dataset $\{(\vec{x}_{i},y_{i})\}_{i=0}^{n-1}$ with
$y_{i}\in\mathbb{R}$. The high dimensional discrete non-uniform Fourier
transform of $\{(\vec{x}_{i},y_{i})\}_{i=0}^{n-1}$ is $\hat{y}_{\vec{k}}=\frac{1}{n}\sum_{i=0}^{n-1}y_{i}\exp\left(-\I 2\pi\vec{k}\cdot\vec{x}_{i}\right)$.
Consider a direction of $\vec{k}$ in the Fourier space, i.e.,
$\vec{k}=k\vec{p}_{1}$, $\vec{p}_{1}$
is a chosen and fixed unit vector. Then
we have $\hat{y}_{k}=\frac{1}{n}\sum_{i=0}^{n-1}y_{i}\exp\left(-\I2\pi(\vec{p}_{1}\cdot\vec{x}_{j})k\right)$,
which is essentially the $1$-d Fourier transform of $\{(x_{\vec{p}_{1},i},y_{i})\}_{i=0}^{n-1}$,
where $x_{\vec{p}_{1},i}=\vec{p}_{1}\cdot\vec{x}_{i}$ is
the projection of $\vec{x}_{i}$ on the direction $\vec{p}_{1}$.  Similarly, one can examine the relative difference between the DNN
output and the target function for the selected important frequencies at each recording step. In the experiments in \citet{xu2019frequency}, $\vec{p}_{1}$
is chosen as the first principle component of the input space. A fully-connected network and a convolutional network are used to learn MNIST and CIFAR10, respectively.  As shown in Fig.~\ref{fig:CMFT}(a) and \ref{fig:CMFT}(c),
low frequencies dominate in both real datasets. As shown in Fig.~\ref{fig:CMFT}(b) and \ref{fig:CMFT}(d), one can
easily observe that DNNs capture low frequencies first and gradually
capture higher frequencies.  
\begin{center}
\begin{figure*}[h]
\begin{centering}
\includegraphics[width=\textwidth]{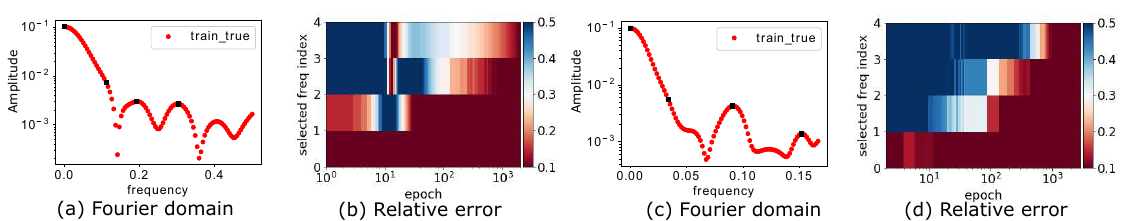} 
\par\end{centering}
\caption{Projection method. (a, b) are for MNIST, (c, d) for CIFAR10. (a, c)
Amplitude $|\hat{y}_{k}|$ vs. frequency. Selected frequencies are
marked by black squares. (b, d) $\Delta_{F}(k)$ vs. training epochs
for the selected frequencies. Reprinted from \citet{xu2019frequency}. \label{fig:CMFT} }
\end{figure*}
\par\end{center}

\subsubsection{Filtering method \label{sec:Filter}}

The projection method examines the F-Principle at only several directions.
To compensate the projection method, one can consider
a coarse-grained filtering method which is able to unravel whether,
in the radially averaged sense, low frequencies converge faster than
high frequencies.
 
% We decompose the frequency space into two domains
The idea of the filtering method is to use a Gaussian filter to derive the low-frequency part of the data and then examine the convergence of the low- and high-frequency parts separately.  The low frequency part can be derived by 
\begin{equation}
    \vec{y}_{i}^{\mathrm{low},\delta}\triangleq(\vec{y}*G^{\delta})_{i},\label{eq:filter-1}
\end{equation}
where $*$ indicates convolution operator, $\delta$ is the standard deviation of the Gaussian kernel. Since the Fourier transform of a Gaussian function is still a Gaussian function but with a standard deviation $1/\delta$, $1/\delta$ can be regarded as a rough frequency width which is kept in the low-frequency part. The high-frequency part can be derived by 
\begin{equation}
    \vec{y}_{i}^{\mathrm{high},\delta}\triangleq\vec{y}_{i}-\vec{y}_{i}^{\mathrm{low},\delta}.\label{eq:yleft-1}
\end{equation}
Then, one can examine 
\begin{equation}
    e_{\mathrm{low}}=\left(\frac{\sum_{i}|\vec{y}_{i}^{\mathrm{low},\delta}-\vec{h}_{i}^{\mathrm{low},\delta}|^{2}}{\sum_{i}|\vec{y}_{i}^{\mathrm{low},\delta}|^{2}}\right)^{\frac{1}{2}},
\end{equation}
\begin{equation}
     e_{\mathrm{high}}=\left(\frac{\sum_{i}|\vec{y}_{i}^{\mathrm{high},\delta}-\vec{h}_{i}^{\mathrm{high},\delta}|^{2}}{\sum_{i}|\vec{y}_{i}^{\mathrm{high},\delta}|^{2}}\right)^{\frac{1}{2}},\label{eq:ehigh}
\end{equation}
where $\vec{h}^{\mathrm{low},\delta}$ and $\vec{h}^{\mathrm{high},\delta}$
are obtained from the DNN output $\vec{h}$.
If $e_{\mathrm{low}}<e_{\mathrm{high}}$ for different $\delta$'s during
the training, F-Principle holds; otherwise, it is falsified. Note the DNN is trained as usual.

As shown in Fig. \ref{fig:Noisefitting-Mnist}, low-frequency part converges faster in the following three settings for different $\delta$'s: a tanh fully-connected network for MNIST, a ReLU shallow convolutional network for CIFAR10, and a VGG16 \citep{simonyan2014very} for CIFAR10.

Another approach to examine the F-Principle in high-dimensional data is to add noise to the training data and examine when the noise is captured by the network \citep{rahaman2018spectral}. Note that this approach contaminates the training data.

\begin{center}
\begin{figure}[h]
\begin{centering}
\subfloat[$\delta=3$, DNN]{\begin{centering}
\includegraphics[width=0.32\textwidth]{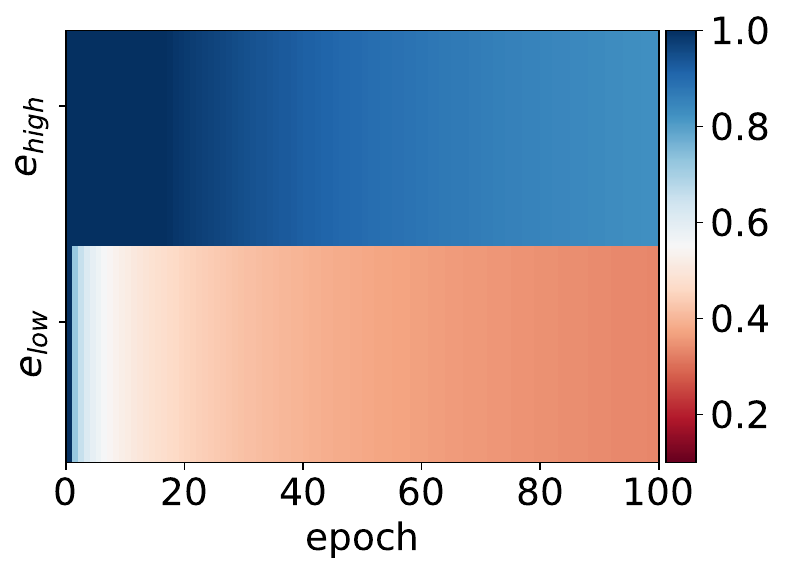} 
\par\end{centering}
}\subfloat[$\delta=3$, CNN]{\begin{centering}
\includegraphics[width=0.32\textwidth]{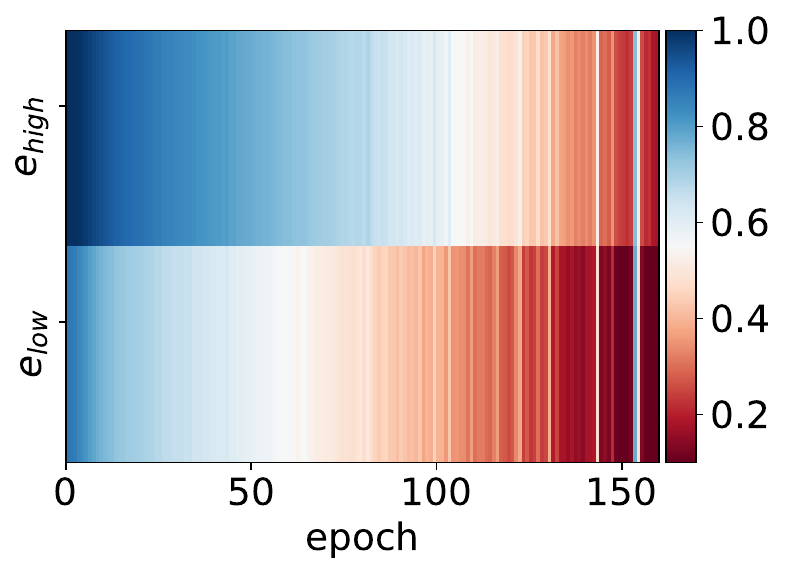} 
\par\end{centering}
}\subfloat[$\delta=7$, VGG]{\begin{centering}
\includegraphics[width=0.32\textwidth]{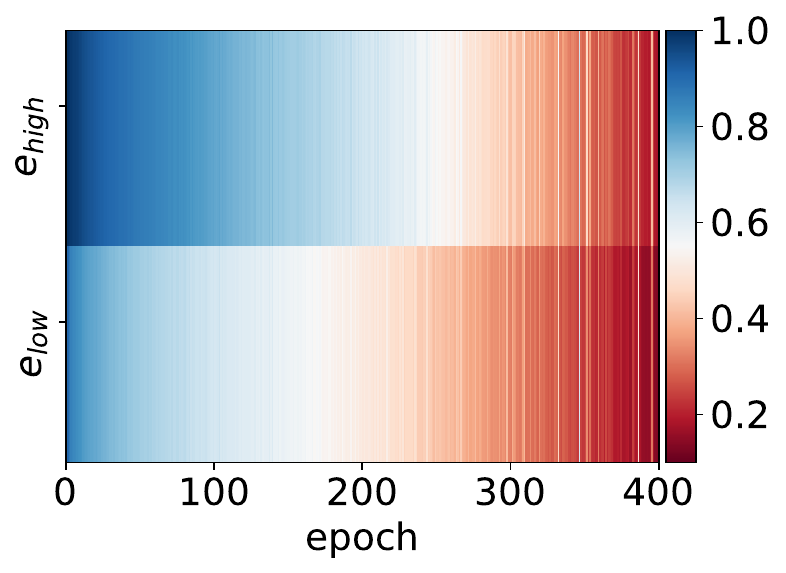} 
\par\end{centering}
}
\par\end{centering}
\begin{centering}
\subfloat[$\delta=7$, DNN]{\begin{centering}
\includegraphics[width=0.32\textwidth]{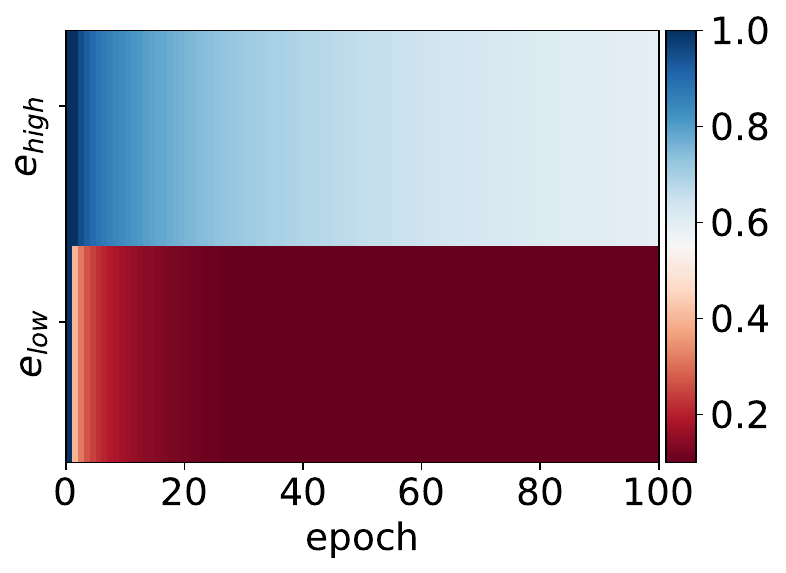} 
\par\end{centering}
}\subfloat[$\delta=7$, CNN]{\begin{centering}
\includegraphics[width=0.32\textwidth]{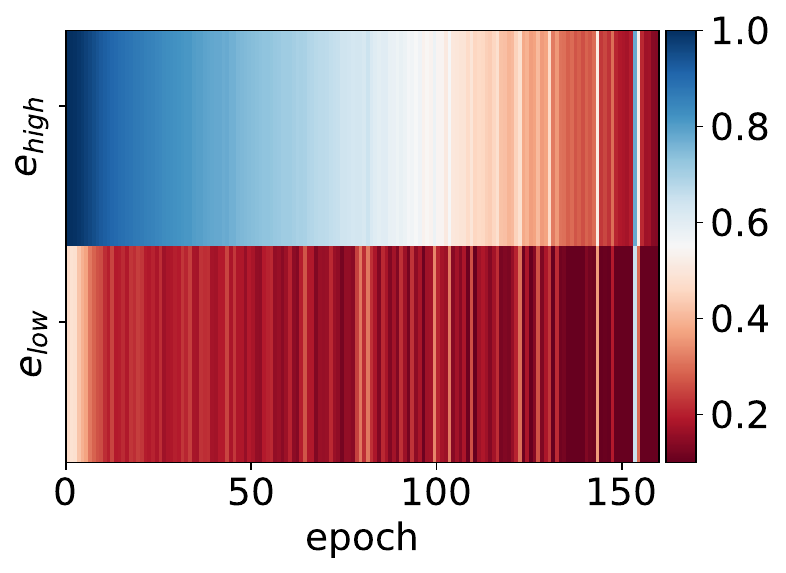} 
\par\end{centering}
}\subfloat[$\delta=10$, VGG]{\begin{centering}
\includegraphics[width=0.32\textwidth]{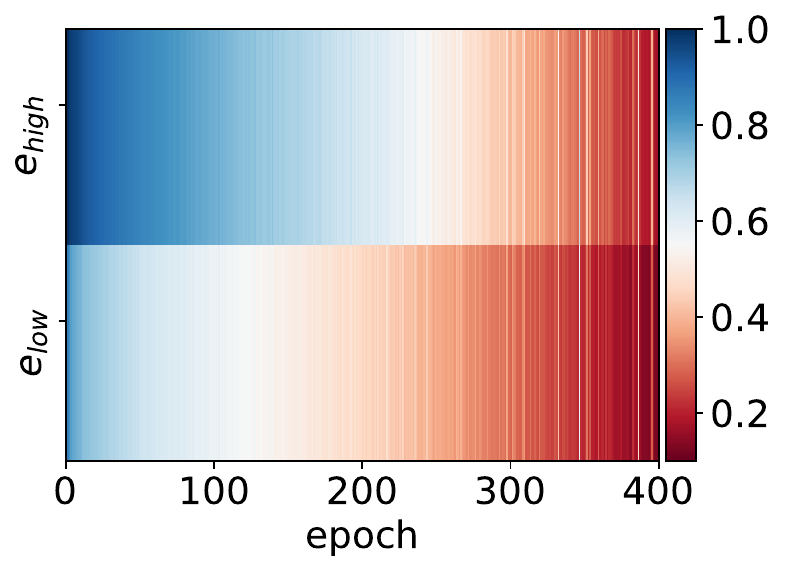} 
\par\end{centering}
}
\par\end{centering}
\caption{F-Principle in real datasets. $e_{\mathrm{low}}$ and $e_{\mathrm{high}}$
indicated by color against training epoch.  Reprinted from \citet{xu2019frequency}. \label{fig:Noisefitting-Mnist} }
\end{figure}
\par\end{center}

%\subsection{Frequency principle beyond gradient-descent-based training}

\section{Theoretical study of F-Principle}

An advantage of studying DNNs from the frequency perspective is that frequency can often be theoretically analyzed. This is especially important in deep learning since  deep learning is often criticized as a black box due to its lack of theoretical support. 
In this section, we review theories of the F-Principle for various
settings. A key mechanism of the F-Principle is based on the regularity of the activation function that is first proposed in \citet{xu2018understanding} and is formally published in \citet{xu2019frequency}.  

The theories have been developed to explore the F-Principle in an ideal setting \citep{xu2019frequency},  in general setting with infinite samples \citep{luo2019theory}, in a continuous viewpoint \citep{e2019machine} and in the regime of wide DNNs (Neural Tangent Kernel (NTK) regime \citep{jacot2018neural}) with specific  sample distributions \citep{cao2019towards,yang2019fine,basri2019convergence,bordelon2020spectrum} or any finite samples \citep{zhang2019explicitizing,zhang2021linear,luo2020theory}.  

\subsection{Ideal setting for analyzing activation function} \label{idealsetting}
% \subsubsection{Ideal setting for analyzing activation function} 
The following presents a simple case to illustrate how the F-Principle  may arise. More details can be found in \citet{xu2018understanding,xu2019frequency}.
The activation function we consider is $\sigma(x)=\tanh(x)$. 
\begin{equation*}
    \sigma(x)=\tanh(x)=\frac{\E^{x}-\E^{-x}}{\E^{x}+\E^{-x}},\quad x\in\mathbb{R}.
\end{equation*}
For a DNN of one hidden layer with $m$ nodes, 1-d input $x$ and
1-d output: 
\begin{equation}
    h(x)=\sum_{j=1}^{m}a_{j}\sigma(w_{j}x+b_{j}),\quad a_{j},w_{j},b_{j}\in{\rm \mathbb{R}},\label{eq: DNNmath}
\end{equation}
where $w_{j}$, $a_{j}$, and $b_{j}$ are training parameters. In the sequel, we will also
use the notation $\theta=\{\theta_{lj}\}$ with $\theta_{1j}=a_{j}$,
$\theta_{2j}=w_{j}$, and $\theta_{3j}=b_{j}$, $j=1,\cdots,m$. Note
that the Fourier transform of $\tanh(x)$ is $\hat{\sigma}(k)=-\frac{\I\pi}{\sinh(\pi k/2)}$.
The Fourier transform of $\sigma(w_{j}x+b_{j})$ with $w_{j},b_{j}\in{\rm \mathbb{R}}$,
$j=1,\cdots,m$ reads as 
\begin{equation}
    \widehat{\sigma(w_{j}\cdot+b_{j})}(k)=\frac{2\pi\I}{|w_{j}|}\exp\Big(\frac{\I b_{j}k}{w_{j}}\Big)\frac{1}{\exp(-\frac{\pi k}{2w_{j}})-\exp(\frac{\pi k}{2w_{j}})}.\label{eq:FSigOri}
\end{equation}
Note that the last term exponentially decays w.r.t. $|k|$. Thus 
\begin{equation}
    \hat{h}(k)=\sum_{j=1}^{m}\frac{2\pi a_{j}\I}{|w_{j}|}\exp\Big(\frac{\I b_{j}k}{w_{j}}\Big)\frac{1}{\exp(-\frac{\pi k}{2w_{j}})-\exp(\frac{\pi k}{2w_{j}})}.\label{eq:FTW}
\end{equation}
Define the amplitude deviation between DNN output and the \emph{target
function} $f(x)$ at frequency $k$ as 
\begin{equation*}
    D(k)\triangleq\hat{h}(k)-\hat{f}(k).
\end{equation*}
Write $D(k)$ as $D(k)=A(k)\E^{\I\phi(k)}$, where $A(k)\in[0,+\infty)$
and $\phi(k)\in\mathbb{R}$ are the amplitude and phase of $D(k)$,
respectively. The loss at frequency $k$ is $L(k)=\frac{1}{2}\left|D(k)\right|^{2}$,
where $|\cdot|$ denotes the norm of a complex number. The total loss
function is defined as: $L=\int_{-\infty}^{+\infty}L(k)\diff{k}$.
Note that according to Parseval's theorem, this loss function
in the Fourier domain is equal to the commonly used loss of mean squared
error, that is, $L=\int_{-\infty}^{+\infty}\frac{1}{2}(h(x)-f(x))^{2}\diff{x}$.

The decrement along any direction, say, with respect to parameter
$\theta_{lj}$, is 
\begin{equation}
    \frac{\partial L}{\partial\theta_{lj}}=\int_{-\infty}^{+\infty}\frac{\partial L(k)}{\partial\theta_{lj}}\diff{k}.\label{eq:GDfreq}
\end{equation}
The absolute contribution from frequency $k$ to this total amount
at $\theta_{lj}$ is 
\begin{equation}
    \left|\frac{\partial L(k)}{\partial\theta_{lj}}\right|\approx A(k)\exp\left(-|\pi k/2w_{j}|\right)F_{lj}(\theta_{j},k),\label{eq:DL2}
\end{equation}
where $\theta_{j}\triangleq\{w_{j},b_{j},a_{j}\}$, $\theta_{lj}\in\theta_{j}$,
$F_{lj}(\theta_{j},k)$ is a function with respect to $\theta_{j}$
and $k$, which is approximate $O(1)$.

When the component at frequency $k$ where $\hat{h}(k)$ is not close
enough to $\hat{f}(k)$, i.e., $A(k)\neq 0$, $\exp\left(-|\pi k/2w_{j}|\right)$ would
dominate $F_{lj}(\theta_{j},k)$ for a small $w_{j}$. Intuitively, the gradient of low-frequency components dominates the training, thus, leading a fast convergence of low-frequency components. \black{If $w_{j}$ is larger, then, the dominance of low frequency is less.} 

\subsection{NTK setting and linear F-Principle} \label{sec:ntkandlfp}

In general, it is difficult to analyze the convergence rate of each frequency due to its high dimensionality and nonlinearity. However, in a linear NTK regime \citep{jacot2018neural}, where the network has a width $m$ approaching infinite and a scaling factor  of $1/\sqrt{m}$, several works have explicitly shown the convergence rate of each frequency.

\subsubsection{NTK dynamics} \label{sec:ntkdynamics}
One can consider the following gradient-descent flow dynamics of the empirical risk $L_{S}$ of a network function $f(\cdot,\vtheta)$ parameterized by $\vtheta$ on a set of training data $\{(\vx_i,y_i)\}_{i=1}^{n}$
\begin{equation}
	\left\{
	\begin{array}{l}
		\dot{\vtheta}=-\nabla_{\vtheta}L_{S}(\vtheta),  \\
		\vtheta(0)=\vtheta_0,
	\end{array}
	\right.
\end{equation}
where
\begin{equation}
	L_{S}(\vtheta)
	= \frac{1}{2}\sum_{i=1}^n(f(\vx_i,\vtheta)-y_i)^2.
\end{equation}
Then the training dynamics of output function $f(\cdot,\vtheta)$ is
\begin{align*}
	\frac{\D}{\D t}f(\vx,\vtheta)
	&= \nabla_{\vtheta}f(\vx,\vtheta)\cdot\dot{\vtheta}\\
	&= -\nabla_{\vtheta}f(\vx,\vtheta)\cdot\nabla_{\vtheta}L_{S}(\vtheta)\\
	&= -\nabla_{\vtheta}f(\vx,\vtheta)\cdot\sum_{i=1}^n \nabla_{\vtheta}f(\vx_i,\vtheta)(f(\vx_i,\vtheta)-y_i)\\
	&= -\sum_{i=1}^n K_m(\vx,\vx_i)(f(\vx_i,\vtheta)-y_i)
\end{align*}
where for time $t$ the NTK evaluated at $(\vx,\vx')\in\Omega\times\Omega$ reads as
\begin{equation}
	K_m(\vx,\vx')(t)=\nabla_{\vtheta}f(\vx,\vtheta(t))\cdot\nabla_{\vtheta}f(\vx',\vtheta(t)).
\end{equation}
\black{At the NTK regime, we denote
\begin{equation}
    K^{*}(\vx,\vx'):=\lim_{m\rightarrow\infty} K_m(\vx,\vx')(t).
\end{equation}
% For simplicity, we ignore $t$ in $K^{*}$ and $\vtheta$. 
The gradient descent of the model thus becomes
\begin{equation}
	\frac{\D}{\D t}\Big(f(\vx,\vtheta(t))-f(\vx)\Big)=-\sum_{i=1}^n K^{*}(\vx,\vx_i)\Big(f(\vx_i,\vtheta(t))-f(\vx_i)\Big).
\end{equation}
Define the residual $\vu(\vx,t)=f(\vx,\vtheta(t))-f(\vx)$. Denote $\mX:=(\vx_1,\cdots,\vx_n)^{T}\in \sR ^{n\times d}$ and $\vY:=(\vy_1,\cdots,\vy_n)^{T}\in \sR ^{n}$ as the training data, $u(\mX,t):=(u(\vx_1,t),\cdots,u(\vx_n,t))\in \sR^{n},\nabla_{\vtheta}f(\mX,\vtheta(t)):=(\nabla_{\vtheta}f(\vx_1,\vtheta(t)),\cdots,\nabla_{\vtheta}f(\vx_n,\vtheta(t)))\in \sR^{n\times M}$ (M is the number of parameters), and denote $K^{*}\in \sR^{n\times n}$ as a matrix with 
\begin{equation}
	K^{*}_{i,j}=K^{*}(\vx_i,\vx_j).  \label{K*}
\end{equation}
Then, one can obtain
\begin{equation}
	\frac{\D u(\mX,t)}{\D t}=- K^{*} u(\mX,t). \label{eq:eigenK}
\end{equation}
In a continuous form, one can define the empirical density $\rho(\vx)=\sum_{i=1}^n\delta(\vx-\vx_i)/n$ and further denote $u_{\rho}(\vx,t)=u(\vx,t)\rho(\vx)$. Therefore, the dynamics for $u$ becomes
\begin{equation}
	\frac{\D}{\D t}u(\vx,t)=-\int_{\sR^d}K^{*}(\vx,\vx')u_{\rho}(\vx',t)
	\diff{\vx'}.\label{eq..DynamicsFiniteWidth}
\end{equation}}
This continuous form renders an integral equation analyzed in \citet{e2019machine}.

The convergence analysis of the dynamics in Eq. (\ref{eq:eigenK}) can be done by performing eigen decomposition of $K^{*}$.  The component in the sub-space of an eigen-vector converges faster with a larger eigen-value.  A series of works further show that the eigenvector with a larger eigen-value is lower-frequency, therefore, \black{providing a rigorous proof for low-frequency bias of DNN training process in NTK regime for two-layer networks.}

Consider a two-layer DNN
\begin{align}
	f(\vx,\vtheta)
	&= \frac{1}{\sqrt{m}}\sum_{j=1}^{m}a_{j}\sigma(\vw_{j}^{\T}\vx+b_{j})\label{eq: 2layer-nn} ,
\end{align}
where the vector of all parameters  $\vtheta$ is formed of the parameters for each neuron ${(a_{j},\vw_{j}^{\T},b_{j})}^{\T}\in\sR^{d+2}$
for $j\in[m]$. 
At the infinite neuron limit $m\to \infty$, the following linearization around initialization 
\begin{equation}
	f^{{\rm lin}}\left(\vx;\vtheta(t)\right)=f\left(\vx;\vtheta(0)\right)+\nabla_{\vtheta}f\left(\vx;\vtheta(0)\right)\left(\vtheta(t)-\vtheta(0)\right) \label{eq:linearNN}
\end{equation}
is an effective approximation of $f\left(\vx;\vtheta(t)\right)$, i.e.,  $f^{{\rm lin}}\left(\vx;\vtheta(t)\right)\approx f\left(\vx;\vtheta(t)\right)$ for any $t$, as demonstrated by both theoretical and empirical studies of neural tangent kernels (NTK) \citep{jacot2018neural,lee_wide_2019}.  Note that,  $f^{{\rm lin}}\left(\vx;\vtheta(t)\right)$, linear in $\vtheta$ and nonlinear in $\vx$, reserves the universal approximation power of $f\left(\vx;\vtheta(t)\right)$ at $m\to \infty$. In the following of this sub-section, we do not distinguish $f\left(\vx;\vtheta(t)\right)$ from $f^{{\rm lin}}\left(\vx;\vtheta(t)\right)$.

%Draw a picture to show the development of F-Principle

\subsubsection{Eigen analysis for two-layer DNN with dense data distribution}
For two-layer ReLU network, $K^{*}$ enjoys good properties for theoretical study. The exact form of $K^{*}$ can be theoretically obtained \citep{xie2017diverse}. \black{Under the condition that training samples distributed uniformly on a sphere, the expectation of the gram matrix for a two-layer ReLU network with respect to the samples can be explicitly computed, then, the spectrum of $K^{*}$ can be obtained through spherical harmonic decomposition \citep{xie2017diverse}. In a rough intuition, each eigen-vector of $K^{*}$ corresponds to a specific frequency. Based on such harmonic analysis, \citet{basri2019convergence,cao2019towards} estimate the eigen values of $K^{*}$, i.e., the convergence rate of each frequency. For an arbitrary data distribution, it is often difficult to obtain the explicit form of the eigen decomposition of $K^{*}$.  \citet{basri2020frequency} further release the condition that data distributed uniformly on a sphere to that data distributed piecewise constant on a sphere but limit the result on 1d sphere.} Similarly under the uniform distribution assumption in NTK regime, \citet{bordelon2020spectrum} show that as the size of the training set grows, ReLU DNNs fit successively higher spectral modes of the target function. Empirical studies also validate that real data often align with the eigen-vectors that have large eigen-values, i.e., low frequency eigen-vectors \citep{dong2019distillation,kopitkov2020neural,baratin2020implicit}.

% \citet{yang2019fine} empirical study the neural network at initialization
% state with inputs on boolean cube, in which each coordinate of the
% input can only be $1$ or $-1$. \citet{yang2019fine} found that
% in such case the neural networks bias towards simple function in the
% sense of eigenfunctions with large eigenvalues. The advantage of considering
% boolean cube, which is different from real data, is that the kernel
% when network width is infinity can be computed for multiple layer.
% They found that the optimal depth of the network that maximizes the
% degree $k$ fractional variance (the largest $k$ eigenvalues) at
% initialization increases with $k$, which may indicate that deeper
% neural network can learn more complex functions. 
 
\subsubsection{Linear F-Principle for two-layer neural network with arbitrary data distribution}\label{sec:lfp}
The condition of dense distribution, such as uniform on a sphere, is often non-realistic in a training. A parallel work \citet{zhang2019explicitizing,zhang2021linear} studies the evolution of each frequency for two-layer wide ReLU networks with any data distribution, including randomly discrete case, and derive the linear F-Principle (LFP) model. \citet{luo2020exact} provides a rigorous version of \citet{zhang2019explicitizing,zhang2021linear} and extend the study of ReLU activation function in \citet{zhang2019explicitizing,zhang2021linear} to general activation functions. \black{\citet{luo2020exact} circumvent the difficulty of eigen decomposition for arbitrary data distribution by analyzing the evolution of the network output in Fourier space. The convergence rate of each frequency can be explicitly obtained, thus, its analysis can be extended to arbitrary data distribution.}

The key idea is to perform Fourier transform of the kernel $K^{*}(\vx,\vx')$ w.r.t. both $\vx$ and $\vx'$. To separate the evolution of each frequency, one has to assume the bias is significantly larger than $1$. In numerical experiments,  by taking the order of bias as the maximum of the input weight and the output weight, one can obtains an accurate approximation of two-layer DNNs with LFP. 

We use a LFP result for two-layer ReLU network for intuitive understanding. Consider  the residual $\vu(\vx,t)=f(\vx,\vtheta(t))-f^{*}(\vx)$, one can obtain
\begin{equation}  \label{eq..lfpoperatorthm.simpleReLU}
    \partial_t\hat{u}=-(\gamma(\vxi))^2 \hat{u_{\rho}}(\vxi), 
\end{equation}
with 
\begin{equation}
    (\gamma(\vxi))^2=\frac{C_1}{\norm{\vxi}^{d+3}} + \frac{C_2}{\norm{\vxi}^{d+1}},
\end{equation}
where $\hat{\cdot}$ is Fourier transform, $C_{1}$ and $C_{2}$ are constants depending on the initial distribution of parameters, $(\cdot)_{\rho}(\vx) := (\cdot)(\vx)\rho(\vx)$. $\rho(\vx)$ is the data distribution, which can be a continuous function or  $\rho(\vx)=\Sigma_{i=1}^{n}\delta(\vx-\vx_i)/n$.

One can further prove that the long-time solution of (\ref{eq..lfpoperatorthm.simpleReLU}) satisfies the following constrained minimization problem
\begin{equation}\label{eq:ndopti}
\begin{aligned}
 \min_{h} &\int \gamma(\vxi)^{-2}|\hat{h}(\vxi)-\hat{h}_{\rm ini}(\vxi)|^2\diff{\vxi},\\
{\rm s.t.} \; &h(\vx_{i})=f^{*}(\vx_i),\;i=1,\ldots,n.
\end{aligned}
\end{equation}
  Based on the equivalent optimization problem in (\ref{eq:ndopti}), each decaying term for 1-d problems ($d=1$) can be analyzed. When $1/\xi^2$ term dominates, the corresponding minimization problem Eq. (\ref{eq:ndopti}) can be rewritten into spatial domain yields 
\begin{equation}\label{eq:LS}
\begin{aligned}
 &\min_{h} \int|h^{'}(x)-h_{\rm ini}^{'}(x)|^2\diff{x},\\
{\rm s.t.} & \quad h(\vx_{i})=f^{*}(\vx_i),i=1,\cdots,n,
\end{aligned}
\end{equation}
where $'$ indicates differentiation. For $h_{\rm ini}(x)=0$, Eq. (\ref{eq:LS}) indicates a linear spline interpolation. Similarly, when $1/\xi^4$ dominates, $\int|h^{''}(x)-h_{\rm ini}^{''}(x)|^2\diff{x}$ is minimized, indicating a cubic spline. In general, the above two power law decay coexist, giving rise to a specific mixture of linear and cubic splines. For high dimensional problems, the model prediction is difficult to interpret because the order of differentiation depends on $d$ and can be fractal. Similar analysis in spatial domain can be found in a subsequent work in \citet{jin2020implicit}.

Inspired by the variational formulation of LFP model in (\ref{eq:ndopti}), \citet{luo2020fourier} propose a new continuum model for the supervised learning. This is a variational problem with a parameter $\alpha>0$:
\begin{align}
& \min_{h\in \fH} Q_{\alpha}[h] =\int_{\sR^d}\langle\vxi\rangle^\alpha\Abs{\hat{h}(\vxi)}^{2}\diff{\vxi},\label{mini-prob} \\
& \mathrm{s.t.}\quad h(\vx_i)=y_i,\quad i=1,\cdots,n,
\end{align}
where $\langle\vxi\rangle=(1+\norm{\vxi}^2)^{\frac{1}{2}}$ is the ``Japanese bracket'' of $\vxi$ and $\fH=\{h(x)|\int_{\sR^d}\langle\vxi\rangle^\alpha\Abs{\hat{h}(\vxi)}^{2}\diff{\vxi}<\infty\}$.

\citet{luo2020fourier} prove that $\alpha=d$ is a critical point. If $\alpha<d$, the variational problem leads to a trivial solution that is only non-zero at the training data point.  If $\alpha>d$,   the variational problem leads to a solution with certain regularity. The LFP model shows that a DNN is a convenient way to implement the variational formulation, which automatically satisfies the well-posedness condition. 

Finally, we give some remarks on the difference between the eigen decomposition and the frequency decomposition. In the non-NTK regime, the eigen decomposition can be similarly analyzed but without informative explicit form. In addition, 
the study of bias from the perspective of eigen decomposition is very limited. For finite networks, which are practically
used, the kernel evolves with training. Thus, it is hard to understand what kind of component converges faster or slower. The eigen mode of the kernel is also difficult to be perceived. In the contrast, frequency decomposition 
is easy to be interpreted, and a natural approach widely used in science.

\black{\subsection{Spectral bias of fully-connected multi-layer networks }}

\citet{luo2019theory} consider fully-connected multi-layer DNNs trained by a general loss function $\tilde{R}_{\fD}(\vtheta)$ and measure the convergence of different frequencies by a mean squared loss. Similarly to the filtering method, the approach in \citet{luo2019theory} is to decompose the frequency domain into a low-frequency part and a high-frequency part. \black{The key idea is as follows. The Fourier spectrum of an activation function with a certain regularity would decay  with a certain rate. This decay rate would lead to that the gradient of the loss function of a particular frequency would decay with frequency.}

Based on the following assumptions, i.e., i) certain regularity of target function, sample distribution function and activation function; ii)  bounded training trajectory with loss convergence. \citet{luo2019theory} prove that the change of high-frequency loss over the total loss decays with the separated frequency with a certain power, which is determined by the regularity assumption. A key ingredient in the proof is that the composition of functions still has a certain regularity, which renders the decay in frequency domain.  This result thus holds for general network structure with multiple layers. Aside its generality, the characterization of the F-Principle is too coarse-grained to differentiate the effect network structure or the speciality of DNNs, but only gives a qualitative understanding.

\citet{e2019machine} present a continuous framework to study machine learning and suggest that the
 gradient flows of neural networks are nice flows, and they obey the frequency principle, basically because they are integral equations. The regularity of integral equations is higher, thus, leading to a faster decaying in the Fourier domain.

\black{\section{Understanding and studying DNN based on F-Principle}}
In this section, we review how F-Principle gains understandings of  over-parameterized DNNs  \citep{zhang2016understanding} and inspires the study of DNNs. Firstly, we use a series of experiments to study what kind of settings can lead to low-frequency bias. Secondly, we utilize F-Principle to qualitatively and quantitatively study the generalization of DNNs. Thirdly, we review the frequency approach for studying DNNs beyond F-Principle. Finally, we review algorithms inspired by F-Principle.
% we show that DNNs, \black{which does not follow F-Principle,} would end up with oscillated output in synthetic example. Thirdly, we show how intuitively the F-Principle explains a strength and a weakness of deep learning \citep{xu2019frequency}. Fourthly, we provide a rationale for the common trick of early-stopping \citep{xu_training_2018}, which is often used to improve the generalization. Fifthly, we estimate an {\it a priori} generalization error bound for wide two-layer ReLU DNN \citep{luo2020exact}. 
% Finally, we discuss the effect of the F-Principle in alleviating the overfitting in Runge's phenomenon.

\black{\subsection{Empirical study of mechanisms underlying spectral bias}}
\subsubsection{Activation}
The analysis in Section \ref{idealsetting} shows the importance of the activation in the F-Principle. For most activations, such as tanh and RelU, they monotonically decay in the frequency domain. \black{Therefore, we can easily observe that DNN training process follows F-Principle, such as the example in Fig. \ref{fig:onelayer}.} We can also design an activation that does not  monotonically decay in the frequency domain but  monotonically increases up to a high frequency, where we expect to observe the frequency convergence order may flip up to a certain frequency. We use Ricker function with parameter $a$,
\begin{equation}
	\frac{1}{15a}\pi^{1/4}\left(1-\left(\frac{x}{a}\right)^2\right)\exp\left(-\frac{1}{2}\left(\frac{x}{a}\right)^2\right).
\end{equation}
With smaller $a$, Ricker function decays from a higher frequency. \black{We perform similar learning task and use the same setting as Fig. \ref{fig:onelayer}.} As shown in Fig. \ref{fig:ricker}, in the first row, when the activation decays from a low frequency, we can clearly observe low frequency converges faster; however, in the second row, when the activation decays from a high frequency, we can not observe any frequency that converges faster, which is consistent with our analysis.

\begin{center}
	\begin{figure}[h]
		\begin{centering}
			\subfloat[]{\begin{centering}
					\includegraphics[width=0.24\textwidth]{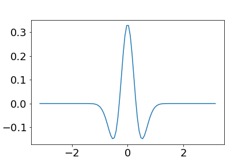} 
					\par\end{centering}
			}\subfloat[]{\begin{centering}
					\includegraphics[width=0.24\textwidth]{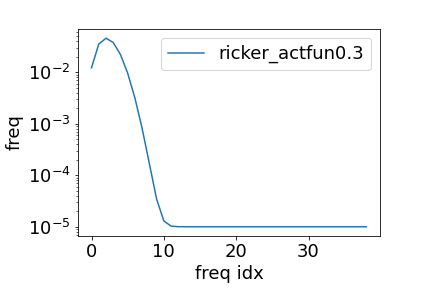} 
					\par\end{centering}
			}\subfloat[]{\begin{centering}
					\includegraphics[width=0.24\textwidth]{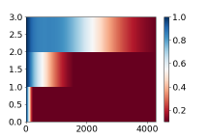} 
					\par\end{centering}
			}
			
			\subfloat[activation in spatial domain]{\begin{centering}
					\includegraphics[width=0.24\textwidth]{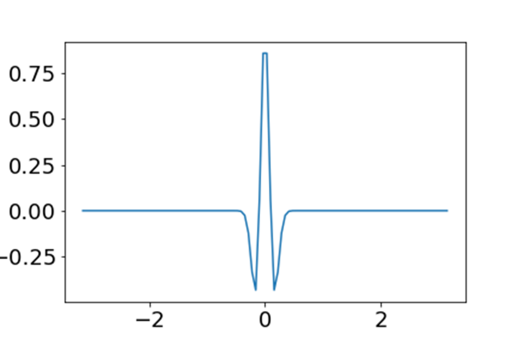} 
					\par\end{centering}
			}\subfloat[activation in frequency domain]{\begin{centering}
					\includegraphics[width=0.24\textwidth]{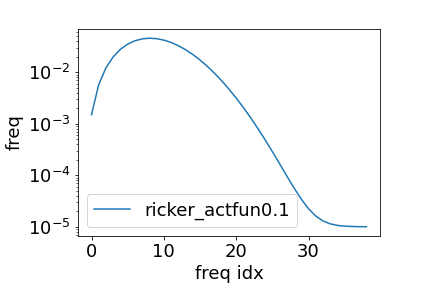} 
					\par\end{centering}
			}\subfloat[convergence order]{\begin{centering}
					\includegraphics[width=0.24\textwidth]{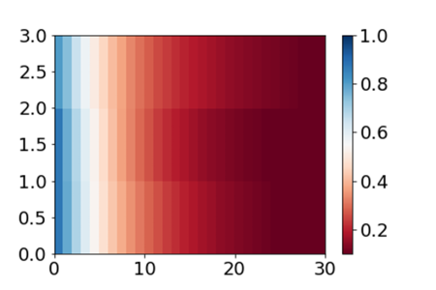} 
					\par\end{centering}
			}\par\end{centering}
		\caption{Ricker activation function. $a=0.3$ for the first row and $a=0.1$ for the second row. \label{fig:ricker} }
	\end{figure}
\par\end{center}

\subsubsection{Frequency weight in the loss function}
The loss function form can affect the frequency convergence. For example, one can explicitly impose a large weight on some specific frequency component to accelerate the convergence of the frequency component. We consider two types of loss functions in learning the function as in Fig. \ref{fig:onelayer}, one is the common mean squared loss $L_{\rm nongrad}$, the other is one with an extra loss of gradient \textcolor{black}{{$L_{\rm grad}$}},
\begin{equation}
	L_{\rm nongrad} = \sum_{i=1}^{n}(f_{\theta}(x_i)-f^{*}(x_i))^{2}/n,
\end{equation}
\begin{equation}
	L_{\rm grad} =L_{\rm nongrad} + \sum_{i=1}^{n}(\nabla_{x} f_{\theta}(x_i)-\nabla_{x}f^{*}(x_i))^{2}/n.
\end{equation}
As shown in Fig. \ref{fig:gradloss}, high frequency converges much faster in the case of the loss with gradient information. The key reason is that, the Fourier transform of $\nabla_{x} f_{\theta}(x_i)$ is the product of the transform of $f_{\theta}(x_i)$ and frequency $\xi$, which is equivalent to adding more priority to higher frequency.

\begin{center}
	\begin{figure}[h]
		\begin{centering}
			\subfloat[$L_{\rm nongrad}$]{\begin{centering}
					\includegraphics[width=0.4\textwidth]{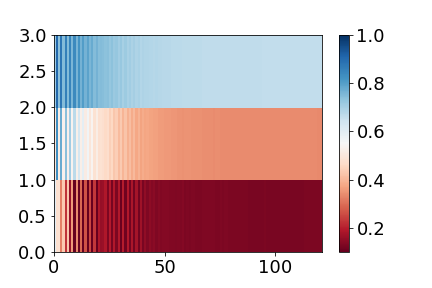} 
					\par\end{centering}
			}\subfloat[$L_{\rm grad}$]{\begin{centering}
					\includegraphics[width=0.4\textwidth]{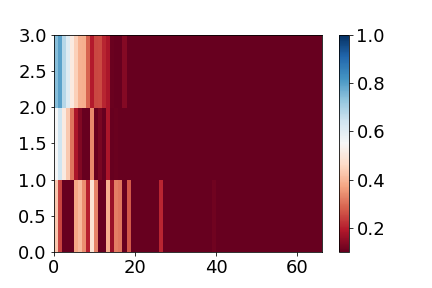} 
					\par\end{centering}
			}\par\end{centering}
		\caption{The frequency convergence for networks with loss function $L_{\rm nongrad}$ and $L_{\rm grad}$. \label{fig:gradloss} }
	\end{figure}
\par\end{center}
\subsubsection{The joint effect of activation and loss}
The analysis in Section \ref{idealsetting} shows that the frequency convergence behavior is the joint effect of activation and loss. A more detailed analysis is shown in Eq. (\ref{eq:jointActLoss}). For common loss functions and activation functions, \black{DNN training process that follows F-Principle can be easily observed.} However, in some specific settings or tasks, such as solving PDEs where the loss function often contains gradient information, \black{F-Principle may not hold.}

\subsection{\black{DNN that violates F-Principle} produces oscillated output}
\black{To examine the utility of the F-Principle, we compare DNNs outputs for two experiments in Fig. \ref{fig:ricker}. Note that the settings for the two examples in Fig. \ref{fig:ricker} are exactly the same except for the hyper-parameter $a$ in the Ricker activation function. For smaller $a$, the output of Ricker activation with small $a$ is more oscillated than that of large $a$. Therefore, the initial output of the network with large $a$, i.e., the one that follows F-Principle, is smooth (Fig. \ref{fig:1dnonfp}(a)), while the one with small $a$, i.e., the one that does not follow F-Principle, is very oscillated (Fig. \ref{fig:1dnonfp}(d)). After training, the network that follows F-Principle learns the training data well in (b), and the DNN output is smooth at test points in (c). However, \black{for the one that does not follow the F-Principle}, the DNN can also learn the training data well in (e), but it is very oscillated at test points in (f). It is also worth to point out that the output of the network is always oscillated at test points during this training. Apparently, such oscillated output usually leads to bad generalization. This example shows the F-Principle is an important factor underlying the good generalization of DNNs.}
\begin{center}
	\begin{figure}[h]
		\begin{centering}
			\subfloat[initial output]{\begin{centering}
					\includegraphics[width=0.3\textwidth]{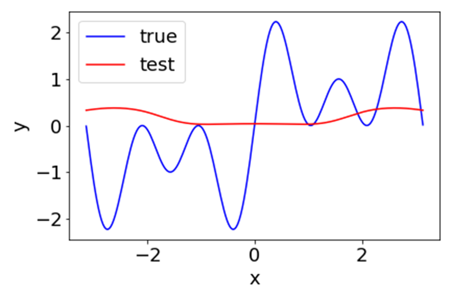} 
					\par\end{centering}}
			\subfloat[final train output]{\begin{centering}
					\includegraphics[width=0.3\textwidth]{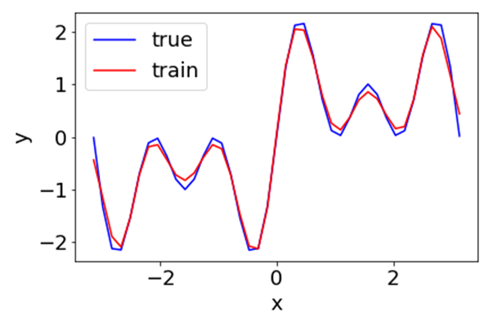} 
					\par\end{centering}
			}\subfloat[final test output]{\begin{centering}
				\includegraphics[width=0.3\textwidth]{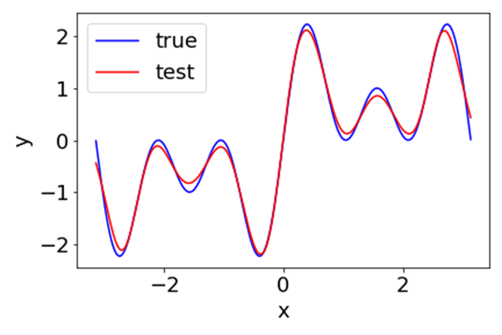} 
				\par\end{centering}
			}
		
			\subfloat[initial output]{\begin{centering}
					\includegraphics[width=0.3\textwidth]{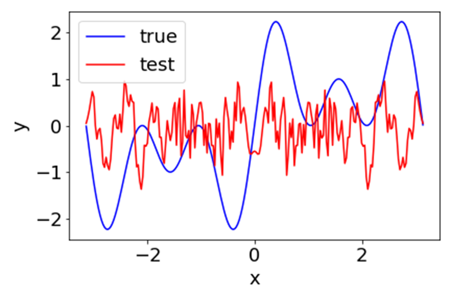} 
					\par\end{centering}
			}	
			\subfloat[final train output]{\begin{centering}
				\includegraphics[width=0.3\textwidth]{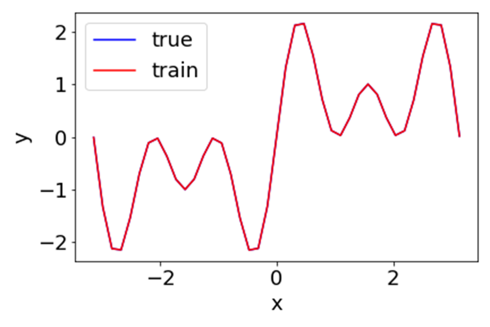} 
				\par\end{centering}
			}\subfloat[final test output]{\begin{centering}
				\includegraphics[width=0.3\textwidth]{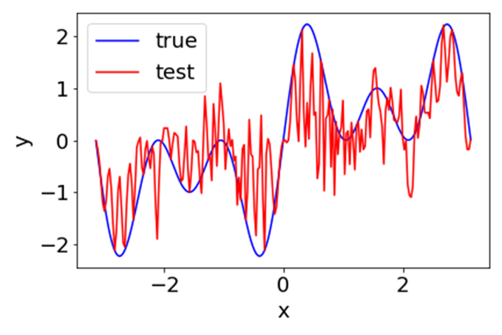} 
				\par\end{centering}
			}\par\end{centering}
		\caption{DNN outputs with Ricker activation function. $a=0.3$ for the first row and $a=0.1$ for the second row. Same experiments as Fig. \ref{fig:ricker}. \label{fig:1dnonfp} }
	\end{figure}
\par\end{center}

\subsection{Strength and weakness}
As demonstrated in the {\it Introduction} part, if the implicit bias or characteristic of an algorithm is consistent with the property of data, the algorithm generalizes well, otherwise not. By identifying the implicit bias of the DNNs in F-Principle, we can have a better understanding of the strength and the weakness of deep learning, as demonstrated by \citet{xu2019frequency} in the following. 

DNNs often generalize well for real problems \cite{zhang2016understanding} but poorly for problems like fitting a parity function \cite{shalev2017failures,nye2018efficient} despite excellent training accuracy for all problems.
The following demonstrates a qualitative difference between these two types of
problems through \emph{Fourier analysis} and use the \emph{F-Principle}
to provide an explanation for different generalization performances of DNNs.

Using the projection method in Sec. \ref{sec:projectmethod}, one can obtain frequencies along the examined directions. For illustration, the Fourier transform of all MNIST/CIFAR10 data along the first principle component
are shown  in Fig.~\ref{fig:parity}(a, b) for MNIST/CIFAR10,
respectively. The Fourier transform of the training data (red dot)  
well overlaps with that of the total data (green) at the dominant
low frequencies. As expected, the Fourier transform of the DNN output with bias of low frequency, evaluated on both training and test data, also overlaps with the true Fourier transform at low-frequency part. Due to the negligible high frequency in these problems, the DNNs generalize well. 

However, DNNs generalize badly for high-frequency functions as follows. For the parity function $f(\vec{x})=\prod_{j=1}^d x_{j}$ defined on
$\Omega=\{-1,1\}^{d}$, its Fourier transform is $\hat{f}(\vec{k})=\frac{1}{2^{d}}\sum_{x\in\Omega}\prod_{j=1}^d x_{j}\E^{-\mathrm{i}2\pi\vec{k}\cdot\vec{x}}=(-\I)^{d}\prod_{j=1}^d\sin2\pi k_{j}$.
Clearly, for $\vec{k}\in[-\frac{1}{4},\frac{1}{4}]^{d}$,
the power of the parity function concentrates at $\vec{k}\in\{-\frac{1}{4},\frac{1}{4}\}^{d}$
and vanishes as $\vec{k}\to\vec{0}$, as illustrated in Fig.~\ref{fig:parity}(c)
for the direction of $\vec{1}_{d}$. Given a randomly sampled
training dataset $S\subset\Omega$ with $s$ points, the nonuniform
Fourier transform on $S$ is computed as $\hat{f}_{S}(\vec{k})=\frac{1}{s}\sum_{x\in S}\prod_{j=1}^d x_{j}\E^{-\I 2\pi\vec{k}\cdot\vec{x}}$.
As shown in Fig.~\ref{fig:parity}(c), $\hat{f}(\vec{k})$ and
$\hat{f}_{S}(\vec{k})$ significantly differ at low frequencies, caused by the well-known \emph{aliasing} effect. Based on the F-Principle,
as demonstrated in Fig.~\ref{fig:parity}(c), these artificial low
frequency components will be first captured to explain the training
samples, whereas the high frequency components will be compromised
by DNN, leading to a bad generalization
performance as observed in experiments.
 
The F-Principle implicates that among all the functions that can
fit the training data, a DNN is implicitly biased during the training
towards a function with more power at low frequencies, which is consistent with the implication of the equivalent optimization problem (\ref{eq:ndopti}). The distribution
of power in Fourier domain of above two types of problems exhibits
significant differences, which results in different generalization
performances of DNNs according to the F-Principle. 
\begin{center}
\begin{figure*}
\begin{centering}
\includegraphics[width=\textwidth]{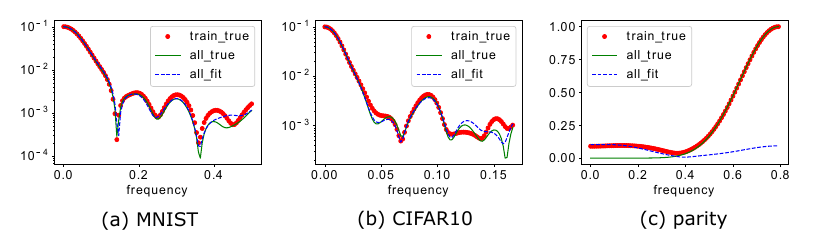} 
\par\end{centering}
\caption{Fourier analysis for different generalization ability. The plot is
the amplitude of the Fourier coefficient against frequency $k$. The
red dots are for the training dataset, the green line is for the whole
dataset, and the blue dashed line is for an output of well-trained
DNN on the input of the whole dataset. For (c), $d=10$.
The training data is $200$ randomly selected points. The settings of (a) and (b) are the same
as the ones in Fig.~\ref{fig:CMFT}. For (c), we use a tanh-DNN with
widths 10-500-100-1, learning rate $0.0005$ under full batch-size
training by Adam optimizer. The parameters
of the DNN are initialized by a Gaussian distribution with mean $0$
and standard deviation $0.05$. Reprinted from \citet{xu2019frequency}. \label{fig:parity}}
\end{figure*}
\par\end{center}

\citet{ma2020slow} show that the F-Principle may be a general mechanism behind the slow deterioration phenomenon in the training of DNNs, where the effect of the ``double descent'' is washed out. 
\citet{sharma2020d} utilize the low-frequency bias of DNNs to study effectiveness of an iris recognition DNN.  
\citet{chen2019muffnet}  show that under the same computational budget, a MuffNet is a better universal approximator
for functions containing high frequency components, thus, better for mobile deep learning.
\citet{zhu2019dspnet} utilize the F-Principle to help understand why high frequency is a limit when DNNs are used to solve spectral deconvolution problem. 
\citet{chakrabarty2019spectral} utilize the idea of F-Principle to study the spectral bias of the deep image prior.

\subsection{Early stopping}

When the training data is contaminated by noise,
early-stopping method is usually applied to avoid overfitting in practice
\citep{lin2016generalization}. By the F-Principle, early-stopping
can help avoid fitting the noisy high-frequency components. Thus,
it naturally leads to a well-generalized solution. \citet{xu_training_2018} use the following
example for illustration.

As shown in Fig. \ref{fig:Generalization}(a), the data are sampled from a function with noise.  The DNN can well fit the sampled training set as the loss
function of the training set decreases to a very small value (green
stars in Fig. \ref{fig:Generalization}(b)). However, the loss function
of the test set first decreases and then increases (red dots in Fig.
\ref{fig:Generalization}(b)). 
Fig. \ref{fig:Generalization}(c), the Fourier transform for the training
data (red) and the test data (black) only overlap around the dominant
low-frequency components. Clearly, the high-frequency components of
the training set are severely contaminated by noise. Around the turning
step --- where the best generalization performance is achieved, indicated
by the green dashed line in Fig.  \ref{fig:Generalization}(b) ---
the DNN output is a smooth function (blue line in Fig.  \ref{fig:Generalization}(a)) in spatial domain and  well captures the dominant peak in frequency domain (Fig.  \ref{fig:Generalization} (c)).
After that, the loss function of the test set increases as
DNN start to capture the higher-frequency noise (red dots in Fig.
\ref{fig:Generalization}b). These phenomena conform with our analysis
that early-stopping can lead to a better generalization performance
of DNNs as it helps avoid fitting the noisy high-frequency components
of the training set. 

As a low-frequency function is more robust w.r.t. input than a high-frequency frequency function, the early-stopping can also enhance the robustness of the DNN. This effect is consistent with the study in \citet{li2020gradient}, which shows that a two-layer DNN, trained only on the input weight and early stopped, can reconstruct the true labels from a noisy data.
\begin{center}
\begin{figure*}
\begin{centering}
\includegraphics[width=\textwidth]{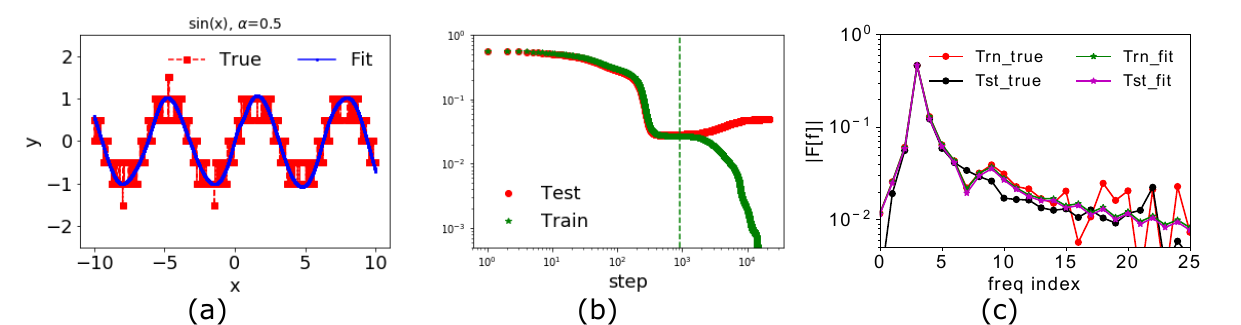}
\par\end{centering}
\caption{Effect of early-stopping on contaminated data. The training set and
the test set consist of $300$ and $6000$ data points evenly sampled
in $[-10,10]$, respectively. (a) The sampled values of the test set
(red square dashed line) and DNN outputs (blue solid line) at the
turning step. (b) Loss functions for training set (green stars) and
test set (red dots) at different recording steps. The green dashed
line is drawn at the turning step, where the best generalization performance
is achieved. (c) The Fourier transform of the true data for the training set (red) and
test set (black), and the Fourier transform of the DNN output for the training set (green),
and test set (magenta) at the turning step. Reprinted from \citet{xu_training_2018}. \label{fig:Generalization}}

\end{figure*}
\par\end{center}

\subsection{Quantitative understanding in NTK regime}
The static minimization problem (\ref{eq:ndopti}) defines an FP-energy $E_{\gamma}(h)=\int \gamma^{-2}|\hat{h}|^2\diff{\vxi}$ that quantifies the preference of the LFP model among all its steady states. Because $\gamma(\vxi)^{-2}$ is an increasing function, say $\gamma(\vxi)^{-2}=\norm{\vxi}^{d+1}$, the FP-energy $\int \norm{\vxi}^{d+1}|\hat{h}|^2\diff{\vxi}$ amplifies the high frequencies while diminishing low frequencies. By minimizing $E_{\gamma}(h)$, problem (\ref{eq:ndopti}) gives rise to a low frequency fitting, instead of an arbitrary one, of training data. By intuition, if target $f^*$ is indeed low frequency dominant, then $h_{\infty}$ likely well approximates $f^*$ at unobserved positions.

To theoretically demonstrate above intuition, \citet{luo2020exact} derive the following, an estimate of the generalization error of $h_{\infty}$ using the {\it a priori} error estimate technique \citep{e_priori_2019}. Because $h(\vx)=f^{*}(\vx)$ is a viable steady state, $E_{\gamma}(h_{\infty})\leq E_{\gamma}(f^*)$ by the minimization problem. Using this constraint on $h_{\infty}$, one can obtain that, with probability of at least $1-\delta$,
\begin{equation}
    \Exp_{\vx} (h_{\infty}(\vx)-f^{*}(\vx))^2
    \leq \frac{E_{\gamma}(f^*)}{\sqrt{n}}C_{\gamma}
    \left(2+4\sqrt{2\log(4/\delta)}\right),\label{generror}
\end{equation}
where $C_{\gamma}$ is a constant depending on $\gamma$. Error reduces with more training data as expected with a decay rate $1/\sqrt{n}$ similar to Monte-Carlo method. Importantly, because $E_{\gamma}(f^*)$ strongly amplifies high frequencies of $f^*$, the more high-frequency components the target function $f^*$ possesses, the worse $h_{\infty}$ may generalize. 

Note that the error estimate is also consistent with another result \citep{arora2019fine} published at similar time. \citet{arora2019fine} prove that the generalization error
of the two-layer ReLU network in NTK regime found by GD is at most
\begin{equation}
    \sqrt{\frac{2Y^{T}(K^{*})^{-1}Y}{n}}, \label{aroraerror}
\end{equation}
where $K^{*}$ is defined in (\ref{K*}), $Y\in \sR^{n}$ is the labels of $n$ training data. If the data $Y$ is dominated more by the component of the eigen-vector that has small eigen-value, then, the above quantity is larger. Since in NTK regime the eigen-vector that has small eigen-value corresponds to a higher frequency, the error bound in (\ref{generror}) is larger, consistent with  (\ref{aroraerror}).

% \subsection{Runge's phenomenon}
% In numerical analysis, Runge's phenomenon is that oscillation at the edges of an interval would occur when we use polynomial interpolation with polynomials of high degree over a set of equispaced interpolation points. Consider using mean squared loss and gradient descent to find the solution of a such polynomial interpolation. Since the solution is unique, the GD with infinite training time would find a same solution that produces Runge's phenomenon. However, similar to the analysis of the F-Principle in the ideal case, apparently, the F-Principle should hold, as shown in Fig. \ref{fig:runge}. This example shows that the F-Principle can co-exist with Runge's phenomenon, a clear over-fitting case. However, as the example of early-stopping shows, the F-Principle can help alleviate the overfitting by early-stopping. Similar analysis for why the double descent is rarely seen in DNN is analyzed in \cite{ma2020slow}.

%\begin{figure}[ht]
%	\begin{centering}
%		\subfloat[]{
%			\includegraphics[width=0.24\textwidth]{pic/runge/e250.png}
%		}
%		\subfloat[]{
%			\includegraphics[width=0.24\textwidth]{pic/runge/e17500.png}
%		}
%		\subfloat[]{
%			\includegraphics[width=0.24\textwidth]{pic/runge/e60750.png}
%		}
%		\subfloat[]{
%			\includegraphics[width=0.24\textwidth]{pic/runge/e128500.png}
%		}
%		\par\end{centering}
%	\caption{Using  mean squared loss and gradient descent to find the solution of a polynomial interpolation with order 11 and 12 equispaced points.}\label{fig:runge}
%\end{figure}

\subsection{Frequency perspective for understanding experimental phenomena}
\paragraph{Compression phase.}
\citet{xu_training_2018} explain the compression phase in information plane, proposed by \citet{shwartz2017opening}, by the F-Principle as follows.  The entropy or information quantifies the possibility of output values, i.e., more possible output values lead to a higher entropy. In learning a discretized function, the DNN first fits
the continuous low-frequency components of the discretized function, i.e., large entropy state.
Then, the DNN output tends to be discretized as the network gradually captures
the high-frequency components, i.e., entropy decreasing. 
Thus, the compression phase appears in the information plane.

\paragraph{Increasing complexity.}
The F-Principle also explains the increasing complexity of DNN output during the training. For common small initialization, the initial output of a DNN is often close to zero. The F-Principle indicates that the output of DNN contains higher and higher frequency during the training. As frequency is a natural concept to characterize the complexity of a function, the F-Principle indicates that the complexity of the DNN output increases during the training.  This increasing complexity of DNN output during training is consistent with previous studies and subsequent works \citep{arpit2017closer,kalimeris2019sgd,goldt2020modeling,he2020assessing,mingard2019neural,jin2019quantifying}. 

% \paragraph{Strength and limitation.}

\paragraph{Deep frequency principle.}
\citet{xu2020deep} propose a deep frequency principle to understand an effect of depth in accelerating the training. For a DNN, the effective target function of the $l$-th hidden layer can be understood in the following way. Its input is the output of the $(l-1)$-th layer. The part from the $l$-th layer is to learn the mapping from the output  of the $(l-1)$-th layer to the true labels. Therefore, the effective target function of the $l$-th hidden layer consists of the output  of the $(l-1)$-th layer and the true labels. \citet{xu2020deep} empirically find a deep frequency principle: The effective target function for a deeper hidden layer biases towards  lower  frequency during the training. Due to the F-Principle, this empirical study provides a rationale for understanding why depth can accelerate the training.

\paragraph{Frequency approach.}
\citet{camuto2020explicit} show that the effect of  Gaussian noise injections to each hidden layer output is equivalent to a penalty of high-frequency components in the Fourier domain.
\citet{rabinowitz2019meta,deng2020meta} use the F-Principle as one of typical
phenomena to study the difference between the normal learning and
the meta-learning. \citet{chen2021frequency} show the frequency principle holds in a broad learning system. \citet{schwarz2021frequency} study the frequency bias of generative models.

\subsection{Inspiring the design of algorithm}  \label{sec:inspirealg} 
In addition to scientific computing reviewed in section \ref{sec:algoforsci}, 
to accelerate the convergence of high-frequency, different approaches are developed in various applications. Some examples are listed in the following.

% PhaseDNN \citep{cai2019phasednn} and MscaleDNN \citep{cai2019multi,liu2020multi} are proposed to solve high frequency PDEs.  

% \citet{jagtap2019adaptive} replace regression in DNNs to solve both forward problems and inverse problems.
% the activation function $\sigma(x)$   by a $\sigma(\mu ax)$,
% where $\mu$ is a fixed scale factor with $\mu\geq1$ and $a$ is
% $a$ is a trainable variable shared for all neurons;

\citet{agarwal2020neural,liang2021reproducing} also design different types of activation functions. \citet{campo2020band} use a frequency filter to help reduce the interdependency between the low frequency and the (harder
to learn) high frequency components of the state-action value approximation to achieve better results in reinforcement learning. Several works use multi-scale input by projecting data into a high dimensional space with a set of sinusoids in efficiently representing complex 3D objects and scenes \citep{tancik2020fourier,mildenhall2020nerf,bi2020neural,pumarola2020d,hennigh2020nvidia,wang2020eigenvector,xu2020positional,hani2020continuous,zheng2020neural,peng2020neural,guo2020object}.
\citet{tancik2020learned} use meta-learning to obtain a good initialization for fast and effective image restoration. Several works utilize frequency-aware information to improve the quality of high-frequency details of images generated by neural networks \citet{chen2020ssd,jiang2020focal,yang2022fregan,li2022spatio}.   
%\citet{jiang2020focal} down-weight low frequencies dynamically in the loss function to generate high-quality images for generative models. 
\citet{xi2020drl} argue that the performance improvement in low-resolution image
classification is affected by the inconsistency of  learning between low-frequency components
and high-frequency components, and \citet{xi2020drl} propose a network structure to overcome this inconsistent issue.

The F-Principle shows that DNNs quickly learn the low-frequency part, which is often dominated in the real data and more robust. At the early stage, the DNN is similar to a linear model \citep{kalimeris2019sgd,hu2020surprising}. Some works take advantage of DNN at early training stage to save training computation cost. The original lottery ticket network \citep{frankle2018lottery} requires a full training of DNN, which has a very high computational cost. Most computation is used to capture high frequency while high frequency may be not important in many cases.  \citet{you2019drawing} show that a small but critical subnetwork emerge at the early training stage (Early-Bird ticket), and the performance of training this small
subnetwork with the same initialization is similar to the training
of the full network, thus, saving significant energy for training
networks. \citet{fu2021cpt,fu2020fractrain} utilize the robust of low frequency by applying low-precision for early training stage to save computational cost without sacrificing the generalization performance.

\black{\section{F-Principle for DNN-based methods for solving PDEs} }\label{sec:scicomp}

Recently, DNN-based approaches have been actively explored for a variety
of scientific computing problems, e.g., solving high-dimensional partial
differential equations \citep{weinan2017deep,khoo2019switchnet,he2018relu,fan2018multiscale,han2019uniformly,e2019model,he2019mgnet,strofer2019data} and molecular dynamics (MD) simulations \citep{han2018deep}. For solving PDEs, one can use DNNs to parameterize the solution of a specific PDE \citep{dissanayake1994neural,weinan2017deep,raissi2019physics,zang2020weak} or the operator of a type of PDE \citep{fan2018multiscale,li2020fourier,lu2021learning,zhang2021mod}. An overview of using DNN to solve high-dimensional PDEs can be found in \citet{han2020algorithms}. We would focus on the former approach.

F-Principle is an important feature of DNN-based algorithms. In this section, we will first review the frequency convergence difference between DNN-based algorithm and conventional methods, i.e., iterative methods and finite element method. Then, we rationalize their difference and understand the F-Principle from the iterative perspective. Finally, we review algorithms developed for overcoming the challenge of high-frequency in DNN-based methods.

 \subsection{Parameterize the solution of a PDE} 
For intuitive illustration, we use Poisson's equation as an example, which has broad applications
in mechanical engineering and theoretical physics \cite{evans2010partial},
 \begin{equation}
-\Delta u(\vx)=g(\vx),\quad \vx \in \Omega,
\end{equation}
with the  boundary condition 
\begin{equation}
u(\vx)=\tilde{g}(\vx),\quad \vx \in \partial \Omega.
\end{equation}
One can use a neural network $u(\vx;\vtheta)$, where $\vtheta$ is the set of DNN parameters. A deep Ritz approach  \cite{weinan2018deep} utilizes the following variational problem
\begin{equation}
u^{*}=\arg\min_{v}J(v),\label{Ritz}
\end{equation}
where the solution of the above minimization problem can be proved to be the solution of the Possion's problem and the energy functional is defined as%
\begin{equation}
J(v) =\int_{\Omega}
\left(  \frac{1}{2}|\nabla v|^{2}+V(\vr)v^{2}\right)  d\vr-
{\displaystyle\int\limits_{\Omega}}
%EndExpansion
g(\mathbf{r})v(\mathbf{r})d\vr
 \triangleq%
%TCIMACRO{\dint \limits_{\Omega}}%
%BeginExpansion
{\displaystyle\int\limits_{\Omega}}
%EndExpansion
E(v(\mathbf{r}))dr.\label{energy}
\end{equation}

\black{In numerical computing, the solution of Poisson problem
is parameterized by a DNN $u(\vx;\vtheta)$, the target functional is discretized in the form of the
first part of Eq. (\ref{ritzlossnum}). The Dirichlet boundary condition is treated as a $L^{2}$ penalty and
discretized as the second part of  Eq. (\ref{ritzlossnum}).} That is,
\begin{equation}
L_{\rm ritz}(\vtheta)=\frac{1}{n}\sum_{\vx\in S}(|\nabla u(\vx;\vtheta)|^2/2-g(\vx)u(\vx;\vtheta))+\frac{\beta}{\tilde{n}}\sum_{\vx\in \tilde{S}}(u(\vx;\vtheta)-\tilde{g}(\vx))^2, \label{ritzlossnum}
\end{equation}
where $S$ is the sample set from $\Omega$ and $n$ is the sample size, $\tilde{n}$  indicates sample set from $\partial\Omega$. The second penalty term with a weight $\beta$ is to enforce the boundary condition. 

A more direct method, also known as physics-informed neural network (PINN) or least squared method, use the following loss function. 
In an alternative approach, one can simply use the loss function of Least Squared  Error (LSE) ,
\begin{equation}
L_{\rm LSE}(\vtheta)=\frac{1}{n}\sum_{\vx\in S}(\Delta u(\vx;\vtheta)+g(\vx))^2+\frac{\beta}{\tilde{n}}\sum_{\vx\in \tilde{S}}(u(\vx;\vtheta)-\tilde{g}(\vx))^2.\label{lselossnum}
\end{equation}
To see the learning accuracy, one can compute the distance between $u(\vx;\vtheta)$ and $u_{\rm true}$,
\begin{equation}
{\rm MSE}(u(\vx;\vtheta),u_{\rm true}(\vx))=\frac{1}{n+\tilde{n}}\sum_{\vx\in S \cup \tilde{S}}(u(\vx;\vtheta)-u_{\rm true}(\vx))^2.
\end{equation}

\subsection{Difference from conventional algorithms}
\subsubsection{Iterative methods}
A stark difference between a DNN-based solver and the Jacobi
method during the training/iteration is that DNNs learn the solution from low- to high-frequencies  \citep{xu2019frequency}, while Jacobi method learns the solution from high- to low-frequencies. Therefore, DNNs would suffer from high-frequency curse.

\paragraph{Jacobi method} \label{sec:jacobi}
Before we show the difference between a DNN-based solver and the Jacobi method, we illustrate the procedure of the Jacobi method. 

Consider a $1$-d Poisson's equation: 
\begin{align}
    & -\Delta u(x)=g(x),\quad x\in\Omega\triangleq(-1,1),\label{eq:Poisson1-1}\\
    & u(-1)=u(1)=0.\label{eq:Poisson1-2}
\end{align}
$[-1,1]$ is uniformly discretized into $n+1$ points with grid size
$h=2/n$. The Poisson's equation in Eq.~(\ref{eq:Poisson1-1}) can
be solved by the central difference scheme, 
\begin{equation}
    -\Delta u_{i}=-\frac{u_{i+1}-2u_{i}+u_{i-1}}{h^{2}}=g(x_{i}),\quad i=1,2,\cdots,n,
\end{equation}
resulting a linear system 
\begin{equation}
    \mat{A}\vec{u}=\vec{g},\label{eq:auf}
\end{equation}
where 
\begin{equation}
    \mat{A}=\left(\begin{array}{cccccc}
    2 & -1 & 0 & 0 & \cdots & 0\\
    -1 & 2 & -1 & 0 & \cdots & 0\\
    0 & -1 & 2 & -1 & \cdots & 0\\
    \vdots & \vdots & \cdots &  &  & \vdots\\
    0 & 0 & \cdots & 0 & -1 & 2
    \end{array}\right)_{(n-1)\times(n-1)}, \label{eq:jacobiA}
\end{equation}

\begin{equation}
    \vec{u}=\left(\begin{array}{c}
    u_{1}\\
    u_{2}\\
    \vdots\\
    u_{n-2}\\
    u_{n-1}
    \end{array}\right),\quad
    \vec{g}=h^{2}\left(\begin{array}{c}
    g_{1}\\
    g_{2}\\
    \vdots\\
    g_{n-2}\\
    g_{n-1}
    \end{array}\right),\quad x_{i}=2\frac{i}{n}.
\end{equation}
A class of methods to solve this linear system is iterative schemes,
for example, the Jacobi method. Let $\mat{A}=\mat{D}-\mat{L}-\mat{U}$, where $\mat{D}$ is the
diagonal of $\mat{A}$, and $\mat{L}$ and $\mat{U}$ are the strictly lower and upper
triangular parts of $-\mat{A}$, respectively. Then, we obtain 
\begin{equation}
    \vec{u}=\mat{D}^{-1}(\mat{L}+\mat{U})\vec{u}+\mat{D}^{-1}\vec{g}.
\end{equation}
At step $t\in\sN$, the Jacobi iteration reads as
\begin{equation}
    \vec{u}^{t+1}=\mat{D}^{-1}(\mat{L}+\mat{U})\vec{u}^{t}+\mat{D}^{-1}\vec{g}.
\end{equation}
We perform the standard error analysis of the above iteration process.
Denote $\vec{u}^{*}$ as the true value obtained by directly performing
inverse of $\mat{A}$ in Eq.~(\ref{eq:auf}). The error at step $t+1$
is $\vec{e}^{t+1}=\vec{u}^{t+1}-\vec{u}^{*}$. Then, $\vec{e}^{t+1}=\mat{R}_{J}\vec{e}^{t}$,
where $\mat{R}_{J}=\mat{D}^{-1}(\mat{L}+\mat{U})$. The converging speed of $\vec{e}^{t}$
is determined by the eigenvalues of $\mat{R}_{J}$, that is, 
\begin{equation}
    \lambda_{k}=\lambda_{k}(\mat{R}_{J})=\cos\frac{k\pi}{n},\quad k=1,2,\cdots,n-1,
\end{equation}
and the corresponding eigenvector $\vec{v}_{k}$'s entry is 
\begin{equation}
    v_{k,i}=\sin\frac{ik\pi}{n},i=1,2,\cdots,n-1.
\end{equation}
So we can write 
\begin{equation}
    \vec{e}^{t}=\sum_{k=1}^{n-1}\alpha_{k}^{t}\vec{v}_{k},
\end{equation}
where $\alpha_{k}^{t}$ can be understood as the magnitude of $\vec{e}^{t}$
in the direction of $\vec{v}_{k}$. Then, 
\begin{equation}
    \vec{e}^{t+1}=\sum_{k=1}^{n-1}\alpha_{k}^{t}\mat{R}_{J}\vec{v}_{k}=\sum_{k=1}^{n-1}\alpha_{k}^{t}\lambda_{k}\vec{v}_{k}.
\end{equation}
\begin{equation*}
    \alpha_{k}^{t+1}=\lambda_{k}\alpha_{k}^{t}.
\end{equation*}
Therefore, the converging rate of $\vec{e}^{t}$ in the direction
of $\vec{v}_{k}$ is controlled by $\lambda_{k}$. Since 
\begin{equation}
    \cos\frac{k\pi}{n}=-\cos\frac{(n-k)\pi}{n},
\end{equation}
the frequencies $k$ and $(n-k)$ are closely related and converge
with the same rate. Consider the frequency $k<n/2$, $\lambda_{k}$
is larger for lower frequency. Therefore, lower frequency converges
more slowly in the Jacobi method.
\paragraph{Numerical experiments}
\citet{xu2019frequency} consider the example with $g(x)=\sin(x)+4\sin(4x)-8\sin(8x)+16\sin(24x)$ such that the exact solution $u_{\mathrm{ref}}(x)$ has several high frequencies.
After training with Ritz loss, the DNN output well matches the analytical solution $u_{\mathrm{ref}}$. For each frequency $k$, we define the relative error as 
$$\Delta_{F}(\vec{k})=|\hat{u}_{\vtheta}(\vec{k})-\hat{u}_{\rm true}(\vec{k})|/\hat{u}_{\rm true}(\vec{k}).$$
Focusing on the convergence of three peaks (inset of Fig.~\ref{fig:Poisson}(a))
in the Fourier transform of $u_{\mathrm{ref}}$, as shown in Fig.~\ref{fig:Poisson}(b),
low frequencies converge faster than high frequencies as predicted
by the F-Principle. For comparison, \citet{xu2019frequency} also use the Jacobi method
to solve problem (\ref{eq:Poisson1-1}). High frequencies converge
faster in the Jacobi method,
as shown in Fig.~\ref{fig:Poisson}(c).

As a demonstration, \citet{xu2019frequency} further propose that DNN can be combined with
conventional numerical schemes to accelerate the convergence of low
frequencies for computational problems. First, \citet{xu2019frequency} solve the Poisson's
equation in Eq.~(\ref{eq:Poisson1-1}) by DNN with $M$ optimization
steps (or epochs). Then, \citet{xu2019frequency} use the Jacobi
method with the new initial data for the further iterations.  A proper choice of $M$ is indicated
by the initial point of orange dashed line, in which low frequencies
are quickly captured by the DNN, followed by fast convergence in high
frequencies of the Jacobi method. Similar idea of using DNN as initial guess for conventional methods is proved to be effective in later works \cite{huang2020int}.

This example illustrates a cautionary tale that, although DNNs have
clear advantage, using DNNs alone may not be the best option because
of its limitation of slow convergence at high frequencies. Taking
advantage of both DNNs and conventional methods to design faster schemes
could be a promising direction in scientific computing problems. 
\begin{center}
\begin{figure*}
\begin{centering}
\includegraphics[width=\textwidth]{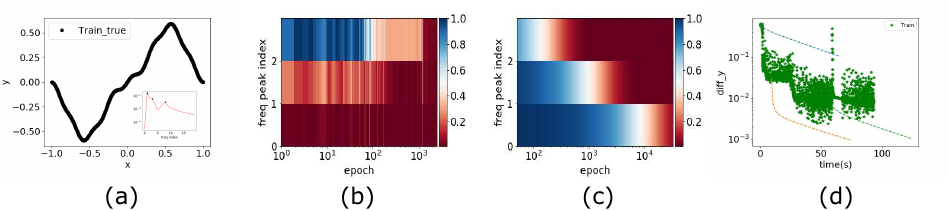} 
\par\end{centering}
\caption{Poisson's equation. (a) $u_{\mathrm{ref}}(x)$. Inset: $|\hat{u}_{\mathrm{ref}}(k)|$
as a function of frequency. Frequencies peaks are marked with black
dots. (b,c) $\Delta_{F}(k)$ computed on the inputs of training data
at different epochs for the selected frequencies for DNN (b) and Jacobi
(c). (d) $\|h-u_{\mathrm{ref}}\|_{\infty}$ at different running time.
Green stars indicate $\|h-u_{\mathrm{ref}}\|_{\infty}$ using DNN alone.
The dashed lines indicate $\|h-u_{\mathrm{ref}}\|_{\infty}$ for the Jacobi
method with different colors indicating initialization by different
timing of DNN training.  \citet{xu2019frequency} use a DNN with widths 1-4000-500-400-1
and full batch training by Adam optimizer \cite{kingma2014adam}.
The learning rate is $0.0005$. $\beta$ is $10$. The parameters
of the DNN are initialized following a Gaussian distribution with
mean $0$ and standard deviation $0.02$. Reprinted from \citet{xu2019frequency}. \label{fig:Poisson} }
\end{figure*}
\par\end{center}
\subsubsection{Ritz-Galerkin (R-G) method}
\citet{wang2020implicit} study the difference between R-G method and DNN methods, reviewed as follows.

\paragraph{R-G method.} We briefly introduce the R-G method \citep{Brenner2008The}. 
For problem \eqref{eq:Poisson1-1}, we construct a functional
\begin{eqnarray}
J(u) = \frac{1}{2}a(u,u)-(g,u),
\end{eqnarray}
where
$$
a(u,v) = \int_\Omega \nabla u(\bm{x})\nabla v(\bm{x}) d\bm{x} ,\quad (g,v)=\int_\Omega g(\bm{x})v(\bm{x}) d\bm{x}. 
$$
The variational form of problem \eqref{eq:Poisson1-1} is the following:
\begin{eqnarray}\label{MPE}
\text{Find}\; u\in H_0^1(\Omega),\; \text{s.t.} \; J(u)=\min_{v\in H_0^1(\Omega)} J(v).
\end{eqnarray}
%the principle of minimal potential energy told us that if $u\in C^2(\Omega)\cap H_0^1(\Omega)$ is the minimum of $J(u)$, then $u$ is the solution of problem \eqref{eq:Poisson1-1}.
The weak form of \eqref{MPE} is to find $u\in H_0^1(\Omega)$ such that  
\begin{eqnarray}\label{VW}
a(u,v)=(g,v), \quad \forall \;v \in H_0^1(\Omega).
\end{eqnarray}
The problem (\ref{eq:Poisson1-1}) is the strong form if the solution $u\in H_0^2(\Omega)$. 
To numerically solve \eqref{VW}, we now introduce the finite dimensional space $U_h$ to approximate the infinite dimensional space $H_0^1(\Omega)$.  Let $U_h \subset H_0^1(\Omega)$ be a subspace with a sequence of basis functions $\{\phi_1,\phi_2,\cdots,\phi_m\}$. The numerical solution $u_h \in U_h$ that we will find can be represented as 
\begin{eqnarray}\label{ucphi}
u_h = \sum_{k=1}^{m} c_k\phi_k,
\end{eqnarray}
where the coefficients $\{c_i\}$ are the unknown values that we need to solve.
Replacing $H_0^1(\Omega)$ by $U_h$, both problems \eqref{MPE} and \eqref{VW} can be transformed to solve the following system:
%R-G method is that find the coefficients $c_i$, such that
\begin{eqnarray}\label{RG}
\sum_{k=1}^{m}c_ka(\phi_k,\phi_j)= (g,\phi_j),\quad j=1,2,\cdots,m.
\end{eqnarray}
From \eqref{RG}, we can calculate $c_i$, and then obtain the numerical solution $u_h$. We usually call \eqref{RG} R-G equation. 

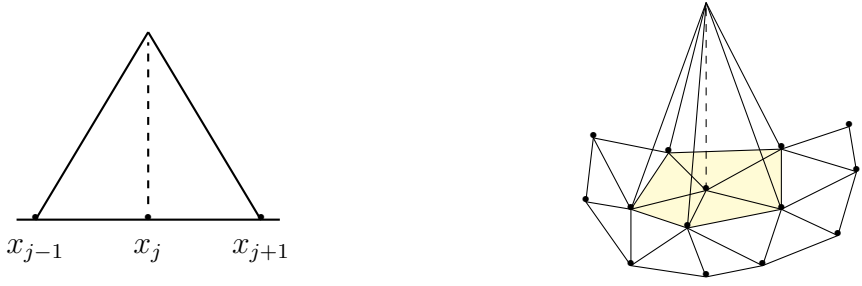
\begin{figure}[h!] 
\centering
\vspace{8pt}
\begin{minipage}[h]{0.45\linewidth}
\begin{tikzpicture}[font=\large,scale=0.5]
\centering
     \tikzset{to/.style={->,>=stealth',line width=.5pt}}
      \node(v1) at (0,0) {$\bigcdot$}; \node[yshift=-0.2cm]  at (v1.south) {$x_{j-1}$};   
      \node(v2) at (3,0) {$\bigcdot$}; \node[yshift=-0.2cm] at (v2.south) {$x_{j}$};  
      \node(v3) at (6,0) {$\bigcdot$}; \node[yshift=-0.2cm]  at (v3.south) {$x_{j+1}$};  
     \node(v4) at (3,5) {};  
      \draw[thick] (-0.5,0.01)--(6.5,0.01); \draw[thick] (0,0)--(3,5); \draw[thick] (6,0)--(3,5);
      \draw[thick, dashed] (v2)--(v4);
\end{tikzpicture}
\end{minipage}
\hspace{0.6cm}
\begin{minipage}[h]{0.45\linewidth}
\begin{tikzpicture}[font=\large,scale=0.5]
\centering
     \tikzset{to/.style={->,>=stealth',line width=.5pt}}
     \coordinate (v1) at (0,0); \coordinate (v7) at (0,5);
     \coordinate (v2) at (-2,-0.5); \coordinate (v3) at (-0.5,-1);
     \coordinate (v4) at (2.,-0.5); \coordinate (v5) at (2,1.1);
     \coordinate (v6) at (-1,1); 
     \draw[fill=yellow!20] (v2)--(v3)--(v4)--(v5)--(v6)--cycle; 
      \node at (v1) {$\bigcdot$};   \node at (v2) {$\bigcdot$}; 
      \node at (v3) {$\bigcdot$};   \node at (v4) {$\bigcdot$};
      \node at (v5) {$\bigcdot$};   \node at (v6) {$\bigcdot$}; 
      \draw (v1)--(v2);  \draw (v1)--(v3);  \draw (v1)--(v4);  \draw (v1)--(v5);  \draw (v1)--(v6); 
      \draw[dashed] (v1)--(v7); \draw (v2)--(v7); \draw (v3)--(v7); \draw (v4)--(v7); \draw (v5)--(v7);\draw (v6)--(v7); 
      \coordinate (e1) at (-3,1.4); \coordinate (e2) at (-3.2,-0.3); \coordinate (e3) at (-2,-2); \coordinate (e4) at (0,-2.3);     
      \coordinate (e5) at (1.5,-2); \coordinate (e6) at (3.5,-1.2); \coordinate (e7) at (4,0.5); \coordinate (e8) at (3.8,1.7);
       \node at (e1) {$\bigcdot$};   \node at (e2) {$\bigcdot$}; 
      \node at (e3) {$\bigcdot$};   \node at (e4) {$\bigcdot$};
      \node at (e5) {$\bigcdot$};   \node at (e6) {$\bigcdot$};   \node at (e7) {$\bigcdot$}; \node at (e8) {$\bigcdot$}; 
      \draw (e1)--(e2)--(e3)--(e4)--(e5)--(e6)--(e7)--(e8); 
      \draw (e1)--(v6);  \draw (e1)--(v2);  \draw (e2)--(v2);  \draw (e3)--(v2);  \draw (e3)--(v3);
      \draw (e4)--(v3);  \draw (e5)--(v3);  \draw (e5)--(v4);  \draw (e6)--(v4);  \draw (e7)--(v4); \draw (e7)--(v5);   
      \draw (e8)--(v5); 
\end{tikzpicture}
\end{minipage}
      \caption{The finite element basis function in 1d and 2d.  Reprinted from \citet{wang2020implicit}.}
      \label{basis}
\vspace{8pt}
 \end{figure}

For different types of basis functions, the R-G method can be divided into finite element method (FEM) and spectral method (SM) and so on. If the basis functions $\{\phi_i(\bm{x})\}$ are local, namely, they are compactly supported, this method is usually taken as the FEM. Assume that $\Omega$ is a polygon, and we divide it into finite element grid $\TT_h$ by simplex, $h=\max_{\tau\in\TT_h}\text{diam}(\tau)$.  A typical finite element basis is the linear hat basis function, satisfying
\begin{eqnarray}
\phi_k(\bm{x_j}) = \delta_{kj},\quad \bm{x_j}\in \N_h,
\end{eqnarray}
where $\N_h$ stands for the set of the nodes of grid $\TT_h$. The schematic diagram of the basis functions in 1-D and 2-D are shown in Fig. \ref{basis}.
%\begin{eqnarray}
%\phi_i(\bm{x}) = \left\{ \begin{array}{cc}
%a_i\cdot \bm{x}+b_i, & \bm{x} \in \Omega_i,\\
%0,  & \text{otherwise},
%\end{array}\right.
%\end{eqnarray}
On the other hand, if we choose the global basis function such as Fourier basis or Legendre basis \citep{Shen2011Spectral}, we call R-G method spectral method.

The error estimate theory of R-G method has been well established. Under suitable assumption on the regularity of solution,  the linear finite element solution $u_h$ has the following error estimate 
$$\Vert u-u_h \Vert_{1} \leq C_1h\Vert u\Vert_{2},$$
where the constant $C_1$ is independent of grid size $h$. The spectral method has the following error estimate 
$$\Vert u-u_h \Vert \leq \frac {C_2}{m^s},$$
where $C_2$ is a constant and the exponent $s$ depends only on the regularity (smoothness) of the solution $u$. If $u$ is smooth enough and satisfies certain boundary conditions, the spectral method has the spectral accuracy. 

\paragraph{Different learning results.} DNNs with ReLU activation function can be proved to be equivalent with a finite element methods in the sense of approximation \citep{he2018relu}. However, the learning results have a stark difference. 
To investigate the difference, we utilize a control experiment, that is,  solving PDEs given $n$ sample points and controlling the number of bases in R-G method and the number of neurons in DNN equal $m$. Although not realistic in the common usage of R-G method, we choose the case $m>n$ because the two methods are completely different in such situation especially when $m\rightarrow \infty$. Then replacing the integral on the r.h.s. of \eqref{RG} with the form of MC integral formula, we obtain 
\begin{eqnarray} \label{VarEqM}
\sum_{k=1}^{m}c_k a(\phi_k,\phi_j)= \frac{1}{n}\sum_{i=1}^n g(\vx_i)\phi_j(\bm{x}_i),\quad j=1,2,\cdots,m.
\end{eqnarray}

\begin{figure}[ht]
    \begin{centering}
    \subfloat[SM]{
        \includegraphics[width=0.24\textwidth]{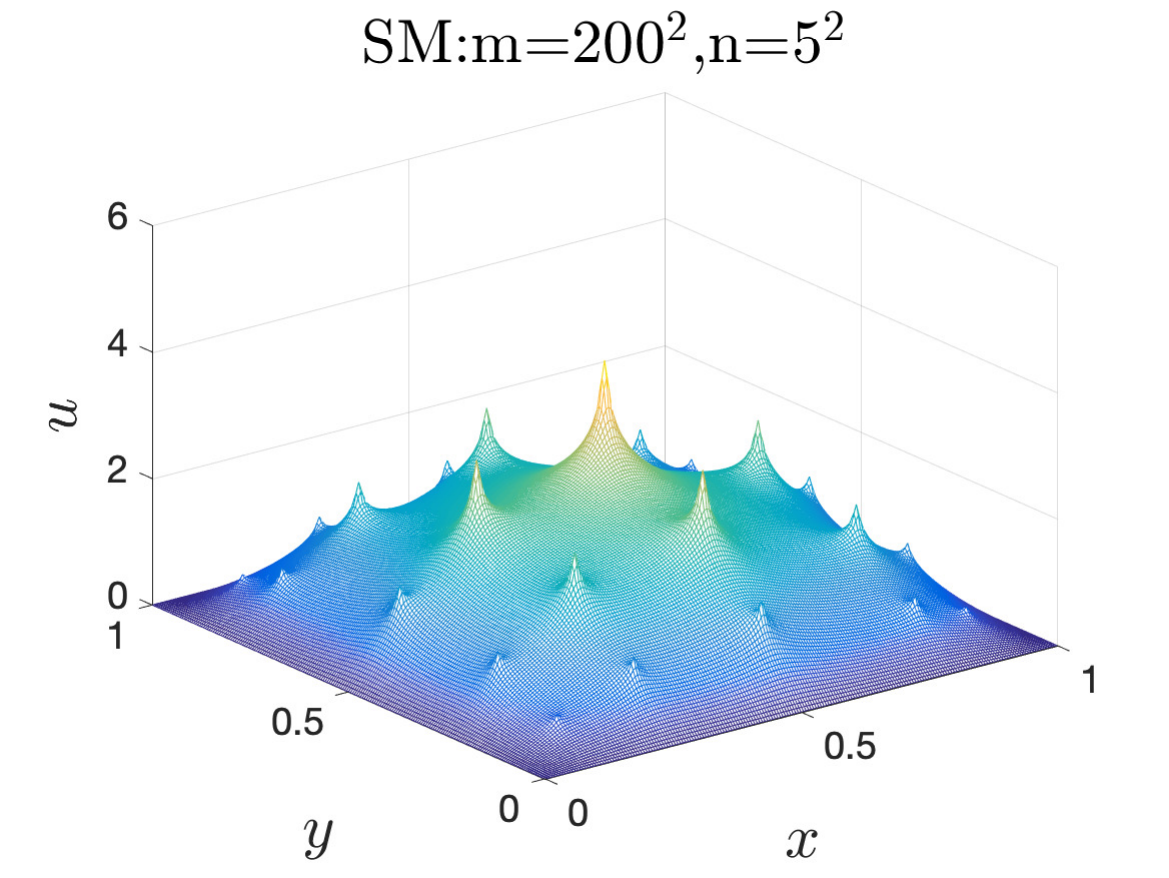}
    }
    \subfloat[FEM]{
        \includegraphics[width=0.24\textwidth]{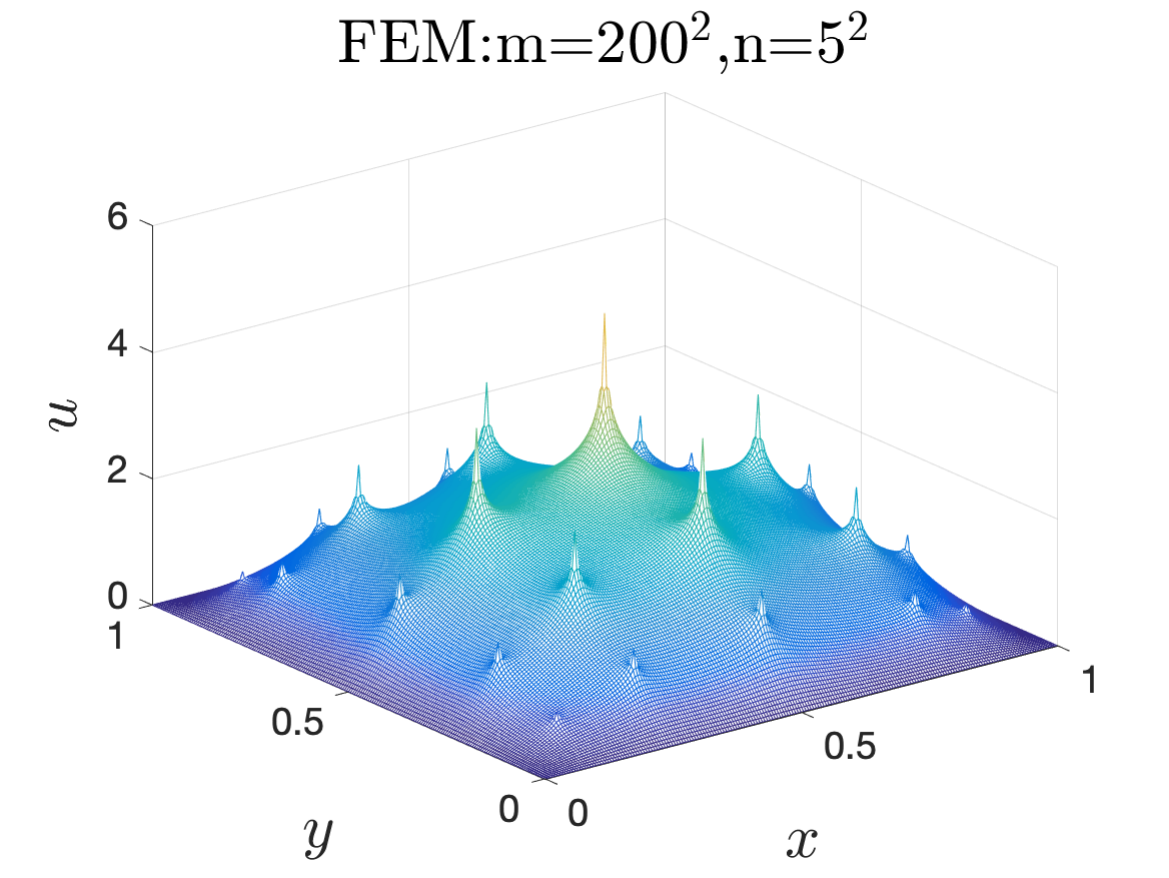}
    }
    \subfloat[ReLU-DNN]{
        \includegraphics[width=0.24\textwidth]{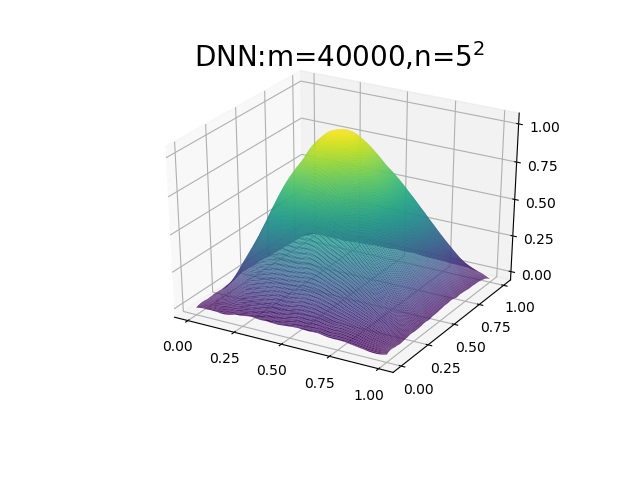}
    }
    \subfloat[Sin-DNN]{
        \includegraphics[width=0.24\textwidth]{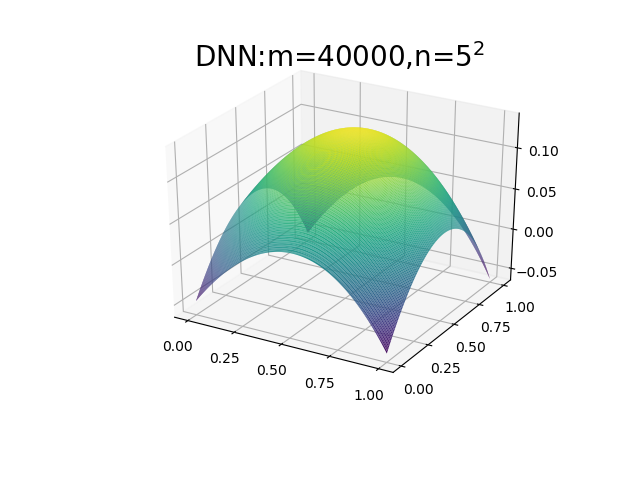}
    }
    \par\end{centering}
    \caption{R-G solutions (a,b) and two-layer DNN solutions (c,d). Reprinted from \citet{wang2020implicit}.}\label{fig:2dfem}
\end{figure}

We  consider the 2-d case 
\begin{eqnarray*}
\left\{\begin{array}{c}
 - \Delta u(\bm{x})=g(\bm{x}),\quad \bm{x}\in(0,1)^2,\\
 u(\bm{x})=0, \quad \bm{x} \in \partial(0,1)^2,
\end{array}
\right.
\end{eqnarray*}
where $\vx=(x, y)$ and we know the values of $g$ at $n=5^2$ points sampled from the function $g(\vx):=g(x, y) = 2\pi^2\sin(\pi x)\sin(\pi y).$
For a large $m$, Fig. \ref{fig:2dfem} (a,b) plot the R-G solutions with Legendre basis and piecewise linear basis function. It can be seen that the numerical solution is a function with strong singularity.  However, Fig. \ref{fig:2dfem}(c,d) show that the two-layer DNN solutions are stable without singularity for large $m$. 
 %We use the Legendre basis in SM and. 
%Due to the limited computing power, the error between the R-G solution and the exact solution is relatively large. 
%It can be seen that the series expression of the reference solution converges, but the convergence speed is slow, and it is difficult for us to draw an accurate image of the reference solution.

The smooth solution of DNN, especially when neuron number is large, can be understood through the low-frequency bias, such as the analysis shown in the LFP theory. This helps understand the wide application of DNN in solving PDEs. For example, the low-frequency bias intuitively explains why DNN solves a shock wave by a smooth solution in \citet{michoski2019solving}.

For R-G method, the following theorem explains why there is singularity in the 2-d case when $m$ is large.

\begin{thm}\label{thm1}
When $m\rightarrow \infty$, the numerical method \eqref{VarEqM} is solving the problem
\begin{eqnarray}\label{Th1Eq}
\left\{\begin{array}{c}
\displaystyle -\Delta u(\bm{x}) = \frac{1}{n}\sum_{i=1}^n \delta(\bm{x}-\bm{x}_i)g(\bm{x}_i), \quad \bm{x}\in \Omega, \\
 u(\bm{x})=0, \quad \bm{x} \in \partial\Omega,
\end{array}
\right.
\end{eqnarray}
where $\delta(x)$ represents the Dirac delta function.
\end{thm}
This theorem shows that if we consider an over-parameterized FEM, it solves the \textcolor{black}{Green's function} of the PDE. In Poisson's problem, the 2-d \textcolor{black}{Green's function} has singularity, thus, leading to singular solution. However, for DNN, due to the F-Principle of low-frequency preference, the solution is always relatively smooth. Although there exists equivalence between DNN-based algorithm and conventional methods for solving PDEs in the sense of approximation, it is important to take the implicit bias when analyzing the learning results of DNNs. 

%  \begin{figure}[ht!]
% \centering
% \begin{minipage}[]{\textwidth}
% 	\includegraphics[width=0.23 \textwidth]{pic/RG/Fig20.eps}\vspace{0.2in}
% 	\includegraphics[width=0.23 \textwidth]{pic/RG/Fig24.eps}
% \end{minipage}
% \caption{\label{2d} (Example 3): R-G solutions with different $m$. The basis functions for the left and the right figures are Legendre basis function and piecewise  linear  basis  function, respectively.  Reprinted from \cite{wang2020implicit}.}
% \end{figure}

%  \begin{figure}[ht!]
% \centering
% \begin{minipage}[]{\textwidth}
% 	\includegraphics[width=0.45\textwidth]{pic/RG/Fig25.eps}\hspace{0.2in} 
% 	\includegraphics[width=0.45\textwidth]{pic/RG/Fig26.eps}
% \end{minipage}
% \caption{\label{2d1} (Example 3): Profile of R-G solutions with different $m$.  Reprinted from \cite{wang2020implicit}.}
% \end{figure}

%  \begin{figure}[ht!]
% \centering
% \begin{minipage}[]{\textwidth}
% 	\includegraphics[width=0.24 \textwidth]{pic/RG/Fig30.png}\vspace{0.2in} 
% 	\includegraphics[width=0.24 \textwidth]{pic/RG/Fig34.png}
% \end{minipage}
% \caption{\label{2dDNN} Two-layer DNN solutions with different $m$. The activation functions  for the left and the right figures are ${\rm ReLU} (x)$ and $\sin (x)$, respectively.  Reprinted from \cite{wang2020implicit}.}
% \end{figure}

\subsection{Understanding F-Principle by comparing the differential operator and the integrator operator}  \label{sec:fpiterative}
In this part, inspired by the analysis in \citet{e2019machine} we use a very non-rigorous derivation to intuitively understand why DNN follows the F-Principle while Jacobi method doesn't and shows a connection between these two methods.

Consider the elliptic equation in $1-d$ :
$$
\left\{\begin{array}{l}
L u:=-\Delta u=g     \quad g \in L^{2}([0,1])\\
u(0)=0 \\
u(1)=0
\end{array} \right.
$$
Suppose that $g$ is sufficiently smooth and thus $u$ is smooth, $N>>1, h=1 / N, x_{i}=i h=\frac{i}{N}, i=0, \ldots, N$.\\
Let $$\bmv=\left(\begin{array}{l}v_{0} \\ v_{1} \\ \vdots\\ v_{N}\end{array}\right)=\left(\begin{array}{l}u(0) \\ u(h) \\ \vdots\\ u(1)\end{array}\right),$$ $v_{0}=v_{N}=0$, we have
$$\left(A_{N} \bmv\right)_{i}=\frac{1}{h^{2}}\left(-v_{i-1}+2 v_{i}-v_{i+1}\right)=g_{i}, i=1, \ldots, N-1$$
$$A_{N} \bmv=\frac{1}{h^{2}}\left(\begin{array}{ccccc}2 & -1 & & & \\ -1 & 2 & -1 & & \\ & -1 & \ddots & \ddots & \\ & & \ddots & \ddots & -1 \\ & & & -1 & 2\end{array}\right)\left(\begin{array}{l}v_{1} \\ v_{2}\\ \vdots \\ v_{N-1}\end{array}\right)$$
$k=1, \cdots, N-1$ (mode) , $i=1, \cdots, N-1$, which can be extended to $i=0, N$ by considering the boundary conditions.

Define $\left(\vw_{k}\right)_{i}:=\sin \frac{i k \pi}{N},  \left(\vec{w}_{k}\right)_{0}:=\left(\vec{w}_{k}\right)_{N}:=0,$ 
$$\begin{aligned}\left(A_{N} \vec{w}_{k}\right)_{i} &=\frac{1}{h^{2}}\left(-\left(\vec{w}_{k}\right)_{i-1}+2\left(\vec{w}_{k}\right)_{i}-\left(\vec{w}_{k}\right)_{i+1}\right) \\ &=\frac{1}{h^{2}}\left(-\sin \frac{(i-1) k \pi}{N}+2 \sin \frac{i k \pi}{N}-\sin \frac{(i+1) k \pi}{N}\right)\\
&=\frac{1}{h^{2}}\left(-2 \sin \frac{i k \pi}{N} \cos \frac{k \pi}{N}+2 \sin \frac{i k \pi}{N}\right) \\
&=\left(\vec{w}_{k}\right)_{i} \frac{2}{h^{2}}\left(1-\cos \frac{k \pi}{N}\right) \\
&=\frac{4}{h^{2}} \sin ^{2} \frac{k \pi}{2 N}\left(\vec{\omega}_{k}\right)_{i} \quad i=1, \cdots, N-1\end{aligned}$$
Thus we have
$$
A_{N} \vec{w}_{k}=\lambda_{k} \vec{w}_{k}, 
$$
where $\lambda_{k}=\frac{4}{h^{2}} \sin ^{2} \frac{k \pi}{N}
k=1, \cdots, N-1$.

As the analysis in Jacobi iteration in Section \ref{sec:jacobi}, for differential operator $A_{N}$, lower frequency converges more slowly. This can also be revealed by optimizing $\bmv$ through a mean square error as follows. 
Define $\ve=\bmv-\vu^{*}$ as the error on the discretized grid points, and
$$
\begin{aligned}
R_{A} :=\frac{1}{2 N}\left\|A_{N} \bmv-g\right\|^{2},
\end{aligned}
$$
By gradient descent flow, we have
$$
\begin{aligned}
\frac{d}{d t} \bmv=&-\nabla_{\bmv} R_{A} \\
=&-\frac{1}{N}A_{N}\left(A_{N}\bmv -\vg\right)
\end{aligned},
$$
then, 
$$
\begin{aligned}
\frac{d}{d t} \ve=-\frac{1}{N}A_{N}^{2} \ve
\end{aligned}.
$$
Similarly, we obtain that low frequency converges more slowly. 

Then, we consider $u(\vx ; \vtheta)$ be the NN function parameterized by $\vtheta$ to approximate the PDE solution. The loss function is similarly defined,
$$
\begin{aligned}
R_{N} :&=\frac{1}{2 N} \sum_{i=1}^{N-1}\left(\left(A_{N} u\right)\left(x_{i}, \vtheta\right)-g\left(x_{i}\right)\right)^{2}+\frac{1}{2} u\left(x_{0}, \vtheta\right)^{2}+\frac{1}{2} u\left(x_{N}, \vtheta\right)^{2} \\
&=\frac{1}{2 N}\left\|A_{N} u-g\right\|^{2}+\frac{1}{2} u\left(x_{0}, \vtheta\right)^{2}+\frac{1}{2} u\left(x_{N}, \vtheta\right)^{2}
\end{aligned},
$$
The gradient flow w.r.t. $\vtheta$ is
$$
\dot{\vtheta}=-\nabla_{\vtheta} R_{N}(\vtheta).
$$
Then, the evolution of $u(\vx,\vtheta)$ is
$$
\begin{aligned}
\frac{d}{d t} u\left(x_{i}, \vtheta\right)=&-\nabla_{\vtheta} u\left(x_{i}, \vtheta\right) \cdot \nabla_{\vtheta} R_N(\vtheta) \\
=&-\nabla_{\vtheta} u\left(x_{i}, \vtheta\right)\left[\frac{1}{N} \sum_{j=1}^{N-1}\left(\left(A_{N}^{2} u\right)_{j}-\left(A_{N} g\right)_{j}\right) \nabla_{\vtheta}u\left(x_{j}, \vtheta\right)\right.\\
&\left.+u\left(x_{0}, \vtheta\right) \nabla_{\vtheta} u\left(x_{0}, \vtheta\right)+u\left(x_{N}, \vtheta\right) \nabla_{\vtheta} u\left(x_{N}, \vtheta\right)\right] \\
=&-\frac{1}{N} \sum_{j=1}^{N-1} K\left(x_{i}, x_{j}\right)\left(\left(A_{N}^{2} u\right)_j-\left(A_{N} g\right)_{j}\right) \\
&-K\left(x_{i}, x_{0}\right) u\left(x_{0}\right)-K\left(x_{i}, x_{N}\right) u\left(x_{N}\right)
\end{aligned},
$$
where 
$K:=\left(K\left(x_{i}, x_{j}\right)\right)_{(N-1)\times(N-1)}=\left(\nabla_{\vtheta} u\left(x_{i}, \vtheta\right) \cdot \nabla_{\vtheta} u\left(x_{j}, \vtheta\right)\right)_{(N-1)\times(N-1)}$ is the spectrum defined in Section \ref{sec:ntkdynamics}. 

We similarly define the error as $e:=u-u^{*}$. Then, 
$$
\begin{aligned}
\frac{d}{d t} e\left(x_{i}, \vtheta\right)=&-\frac{1}{N}\left[\sum_{i=1}^{N-1} K\left(x_{i}, x_{j}\right)\left(A_{N}^{2} e\right)_{j}\right.\\
&\left.-N K\left(x_{i}, x_{0}\right) e_{0}-N K\left(x_{i}, x_{N}\right) e_{N}\right]
\end{aligned}
$$
where $K=\left(K\left(x_{i}, x_{j}\right)\right)
, i, j=0, \cdots, N.  $
% $$A_{N} \bmv=\frac{1}{h^{2}}\left(\begin{array}{ccccc}2 & -1 & & & \\ -1 & 2 & -1 & & \\ & -1 & \ddots & \ddots & \\ & & \ddots & \ddots & -1 \\ & & & -1 & 2\end{array}\right)\left(\begin{array}{l}v_{1} \\ v_{2}\\ \vdots \\ v_{N-1}\end{array}\right), $$
Define $\bar{e}=(e_{0},\cdots,e_{N})$ and consider the augmented matrix:

$$\bar{A}_{N} \bar{\bmv}=\frac{1}{h^{2}}\left(\begin{array}{ccccccc}C\\ & 2 & -1 & & & \\ & -1 & 2 & -1 & & \\ & & -1 & \ddots & \ddots & \\ & & & \ddots & \ddots & -1 \\ & & & & -1 & 2\\ & & & &  & &C\end{array}\right)\left(\begin{array}{l}v_{0} \\ v_{1}\\ \vdots \\ v_{N}\end{array}\right). $$
where $C$ is to be determined. For $j=1, \cdots, N-1$, $\left.\bar{A}_{N}^{2} \bar{\ve}\right|_{j}=\left.A_{N}^{2} \ve \right|_{j}$, for $j=0$,
$$
\begin{aligned}
&\left.\bar{A}_{N}^{2} \bar{e}\right|_{0}=\left(\frac{C}{h^{2}}\right)^{2} e_{0}=N e_{0} \Leftrightarrow \frac{C}{h^{2}}=\sqrt{N} \\
&C=h^{2} \frac{1}{\sqrt{h}}=h^{3 / 2}
\end{aligned}
$$
Taken together, we have
\begin{equation}
	\frac{d}{d t} \bar{e}=-\frac{1}{N} K \bar{A}_{N}^{2} \bar{e}. \label{eq:jointActLoss}
\end{equation}

% We consider solving a problem
% \begin{equation}
% 	\fL u(\vx)=g(\vx).
% \end{equation}
% The empirical risk $L_{S}$ of a network function $u(\cdot,\vtheta)$ parameterized by $\vtheta$ on a set of training data $\{\vx_i\}_{i=1}^{n}$ is
% \begin{equation}
% 	L_{S}(\vtheta)
% 	= \frac{1}{2}\sum_{i=1}^n(\fL u(\vx_i,\vtheta)-g(\vx_{i}))^2.
% \end{equation}
% Similar to Section \ref{sec:ntkdynamics},  the training dynamics of output function $u(\cdot,\vtheta)$ is
% \begin{align*}
% 	\frac{\D}{\D t}u(\vx,\vtheta)
% 	&= \nabla_{\vtheta}u(\vx,\vtheta)\cdot\dot{\vtheta}\\
% 	&= -\nabla_{\vtheta}u(\vx,\vtheta)\cdot\nabla_{\vtheta}L_{S}(\vtheta)\\
% 	&= -\nabla_{\vtheta}u(\vx,\vtheta)\cdot\sum_{i=1}^n \nabla_{\vtheta} \fL u(\vx_i,\vtheta)(\fL u(\vx_i,\vtheta)-g(\vx_{i}))
% \end{align*}

% In the discretized case,
% \begin{align*}
% 	\vu(\vtheta,t+1)
% 	&\sim \vu(\vtheta,t) -K_m A^{T}(A \vu(\vtheta,t)-\vg)\\
% 	&=(I - K_m A^{T}A) \vu(\vtheta,t)-K_m A^{T}\vg
% \end{align*}
% where $A$ is the discretized operator $\fL$, such as the differential case in Eq. \ref{eq:jacobiA}, $K_m$ is the spectrum defined in Section \ref{sec:ntkdynamics}. 

Although for differential operator (derivative w.r.t. input), lower-frequency mode converges more slowly, for integral operator (loss consists of the summation w.r.t. input), i.e., spectrum $K$, lower-frequency mode (see Section \ref{sec:ntkandlfp}) converges faster. Therefore, there is a competition between $K$ and $A_{N}$, which is the competition between integral and differential operators (also see analysis in \citet{e2019machine}). The effect of neural network can also be understood as a preconditioner. Another easy way to understand why differential operator enables faster convergence of high frequency is that the Fourier transform of $\nabla u(x)$ is $\xi \hat{u}(\xi)$, that is, a higher frequency would have a higher weight in the loss function.

\subsection{Algorithm design to overcome the challenge of high-frequency } \label{sec:algoforsci}
The F-Principle provides valuable theoretical insights of the limit of DNN-based algorithms, that is, the challenge of high-frequency \citep{xu2019frequency}. 

To overcome the challenge of high-frequency in DNN-based algorithms, a series of methods are proposed. Some approaches are reviewed as follows. 

\textbf{Learning Fourier coefficients.} PhaseDNN \cite{cai2019phasednn} convert high frequency component of the data downward to a low frequency spectrum for learning, and then convert the learned one back to the original high frequency. Another way to understand PhaseDNN is to expand the target function by a Fourier series, and neural networks are used to learn the coefficients.  \citet{peng2020accelerating} call such method as Prior Dictionary based Physics-Informed Neural Networks (PD-PINNs). However, due to the fact that number of Fourier terms exponentially increases with the dimension, the PhaseDNN would suffer from the curse of dimensionality.

\begin{figure}[htbp]
	\centering
	\subfloat[MscaleDNN-1]{\includegraphics[width=0.45\linewidth]{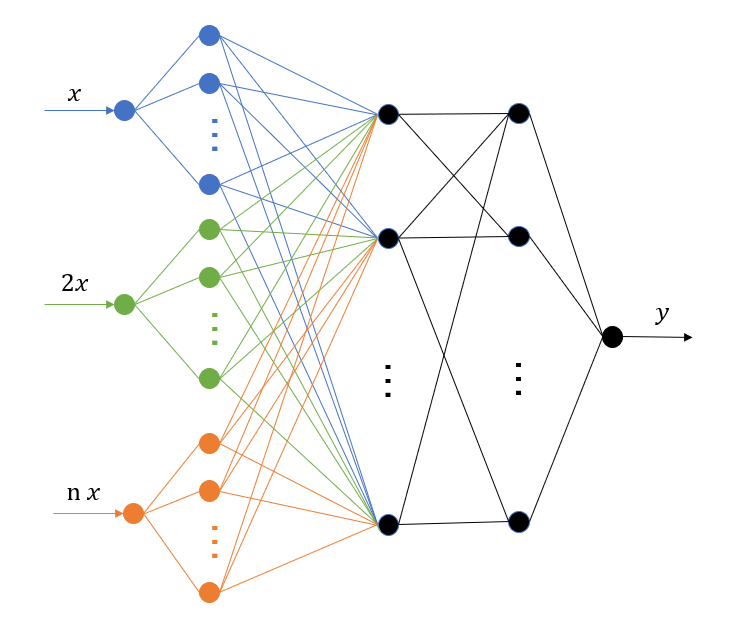}}
	\subfloat[MscaleDNN-2]{\includegraphics[width=0.45\linewidth]{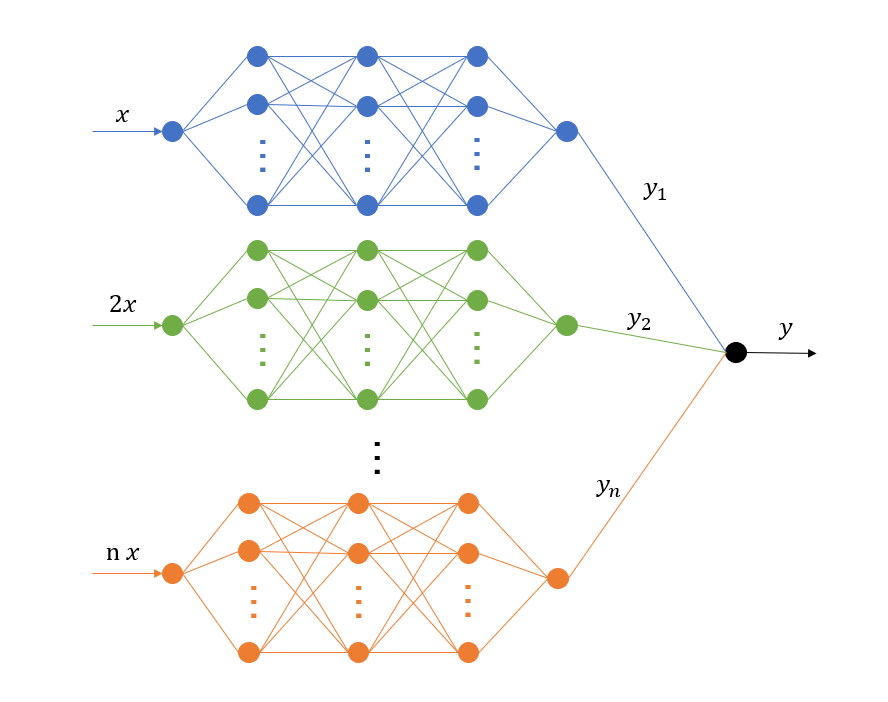}}
	\caption{Illustration of two MscaleDNN structures.   Reprinted from \citet{liu2020multi}.}
	\label{net}
\end{figure}

\textbf{Multi-scale DNN.}  To alleviate the high-frequency difficulty for high-dimensional problem, a Multi-scale DNN (MscaleDNN) method, originally proposed in \citet{cai2019multi} and completed in \citet{liu2020multi}, considers the frequency conversion only in the radial direction. The conversion in the frequency space can be done by a scaling, which is equivalent to an inverse scaling in the spatial space. Therefore, we can use the following ansatz to fit high-frequency data
\begin{equation}
	f(\vx)\sim 
	{\displaystyle\sum\limits_{i=1}^{M}} 
	f_{\theta^{n_{i}}}(\alpha_{i}\vx).\label{f_app}
\end{equation}
This can be easily implemented by multiplying the input to different neurons in the first hidden layer with different constant scalings. Fig. \ref{net} shows two examples of MscaleDNN structures. 

\noindent {\bf MscaleDNN-1} For the first kind,  a MscaleDNN takes the following form
\begin{equation}
    f_{\vtheta}(\vx) = \vW^{[L-1]} \sigma\circ(\cdots (\mW^{[1]} \sigma\circ(\vK\odot(\mW^{[0]} \vx) + \vb^{[0]} ) + \vb^{[1]} )\cdots)+\vb^{[L-1]}, \label{mscalednn}
\end{equation}
where $\vx\in\mathbb{R}^d$, $\mW^{[l]}\in\mathbb{R}^{m_{l+1}\times m_{l}}$, $m_l$ is the neuron number of $l$-th hidden layer, $m_0=d$, $\vb^{[l]}\in\mathbb{R}^{m_{l+1}}$,
$\sigma$ is a scalar function and ``$\circ$'' means entry-wise operation, $\odot$ is the Hadamard product and 
\begin{equation}
\vK=(\underbrace{a_1,a_1,\cdots,a_1}_{\text{1st part}},a_2,\cdots,a_{i-1},\underbrace{a_i,a_i,\cdots,a_i}_{\text{ith part}},\cdots,\underbrace{a_{N},a_{N}\cdots,a_{N}}_{\text{Nth part}})^T,
\end{equation}
where $\vK\in\mathbb{R}^{m_{1}}$, $a_i=i$ or $a_i=2^{i-1}$. This structure is called Multi-scale DNN-1 (MscaleDNN-1).
\bigskip

\noindent {\bf MscaleDNN-2} A second kind of multi-scale DNN is given in Fig. \ref{net}(b), as a sum of $N$ subnetworks, in which each scale input goes through a subnetwork.  In MscaleDNN-2, weight matrices from $W^{[1]}$ to $W^{[L-1]}$ are block diagonal. Again, the scale coefficient $a_i=i$ or $a_i=2^{i-1}$.

\citet{wang2020multi,li2020multi,li2021subspace} further use MscaleDNN to solve more multi-scale problems. Another key factor in practical experiments, without too much understanding, is the effect of activation function. 
\citet{liu2020multi} use activation functions with compact support. \citet{huang2021solving,li2021subspace} adopt the MscaleDNN structure and use the sine and cosine functions as the activation function with different scalings for neurons in the first hidden layer, obtaining better results in solving PDEs.

\textbf{Fourier feature network.} \citet{tancik2020fourier} map input $\vx$ to $$\gamma(\vx)=[a_1 \cos(2\pi \vb_{1}^{T}\vx), a_1 \cos(2\pi \vb_{1}^{T}\vx),\cdots ,a_m \cos(2\pi \vb_{m}^{T}\vx),a_m \cos(2\pi \vb_{m}^{T}\vx)]$$ for imaging reconstruction tasks. $\gamma(\vx)$ is then used as the input to neural network. \citet{wang2021eigenvector} extend the Fourier feature network for PDE problem, where the selection for $b_i$ is from different ranges. \citet{mildenhall2021nerf} successfully apply this multiscale Fourier feature input in the neural radiance fields for view synthesis.

\textbf{Adaptive activation functions.} \citet{jagtap2019adaptive} replace the activation function $\sigma(x)$   by a $\sigma(\mu ax)$, where $\mu$ is a fixed scale factor with $\mu\geq1$ and $a$ is a trainable variable shared for all neurons. \citet{liang2021reproducing}  employ several basic functions and their learnable linear combination to construct neuron-wise data-driven activation functions.

\textbf{Large weight for high frequency.} \citet{biland2019frequency}  explicitly impose high frequencies with higher priority in the loss function to accelerate the simulation of fluid dynamics.  However, Fourier transform for high-dimensional function is computational \textcolor{black}{costly}.

\section{Anti-F-Principle}
F-Principle is rather common in training DNNs. As we have understood F-Principle to a certain extent, it is also easy to construct examples in which F-Principle does not hold, i.e., anti-F-Principle. As analyzed in Section \ref{sec:lfp}, if the priority of high frequency is too high, the optimization problem would lead to trivial solutions. Examples in overparameterized finite element method in Fig. \ref{fig:2dfem} is also an example. In this section, we review some anti-F-Principle examples.

\subsection{Derivative w.r.t. input}
Imposing high priority on high frequency can alleviate the effect of the F-Principle and sometimes \black{a phenomenon that follows anti-F-Principle} can be observed. Similar to Section \ref{sec:fpiterative}, if the loss function contains the gradient of the DNN output w.r.t. the input, it is equivalent to impose higher frequency with higher weight in the loss function. Then, whether there exists a F-Principle depends on the competition between the activation regularity and the loss function. If the loss function endorses more priority for the high frequency to compensate the low-priority induced by the activation function, an anti-F-Principle emerges. Since gradient often exists in solving PDEs, \black{the anti-F-Principle can hold in solving a PDE} by designing a loss with high-order derivatives. Some analysis and numerical experiments can also be found in \citet{lu2019deepxde,e2019machine}. 

\subsection{Large weights}

Another way to observe \black{phenomenon that follows anti-F-Principle} is using large values for network weights. As shown in the analysis of ideal setting in Sec. \ref{idealsetting}, large weights alleviate the dominance of low-frequency in Eq. (\ref{eq:DL2}). In addition, large values would also cause large fluctuation of DNN output at initialization (experiments can be seen in \citet{xu_training_2018}), the amplitude term in Eq. (\ref{eq:DL2}) may endorse high frequency  larger priority, leading to an anti-F-Principle, which is also studied in \citet{yang2019fine}. In the NTK regime \citep{jacot2018neural}, \citet{zhang2019type} theoretically show that the fluctuation of the initial output would be kept in the learned function after training.

\section{Conclusion}

F-Principle is very general and important for training DNNs. It serves as a basic principle to understand DNNs and to inspire the design of DNNs. As a good starting point, F-Principle leads to more interesting studies for better understanding DNNs. For examples, Empirical study also finds the F-Principle holds in non-gradient training process \cite{ma2021frequency}. It remains unclear how to build a theory of F-Principle for general DNNs with arbitrary sample distribution and how to study the generalization error. The precise description of F-Principle is only done in the NTK regime \citep{jacot2018neural}. It is not clear whether it is possible to obtain a similarly precise description in the mean-field regime described by partial differential equations \citep{mei2018mean,sirignano_mean_2020,rotskoff_parameters_2018}.
The Fourier analysis can be used to study DNNs from other perspectives, such as the effect of different image frequency on the learning results.

As a general implicit bias, F-Principle is insufficient to characterize the exact details of the training process of DNNs beyond NTK. To study the nonlinear behavior of DNNs in detail, it is important to study DNNs from other perspectives, such as the loss landscape, the effect of width and depth, the effect of initialization, etc. For example, \citet{zhang2019type,luo2020phase} have studied how initialization affects the implicit bias of DNNs  and \citet{luo2020phase} draw a phase diagram for wide two-layer ReLU DNNs \citep{luo2020phase}. \citet{Zhang2019embedding,zhang2021embedding} show an embedding principle that the loss landscape of a DNN ``contains'' all the critical points of all the narrower DNNs.
\section*{Acknowledgements}
This work is sponsored by the National Key R\&D Program of China  Grant No. 2022YFA1008200 (Z. X., Y. Z., T. L.), the National Natural Science Foundation of China Grant No. 92270001 (Z. X.), 12371511 (Z. X.), 12101402 (Y. Z.), 12101401 (T. L.),  the Lingang Laboratory Grant No.LG-QS-202202-08 (Y.Z.),  Shanghai Municipal Science and Technology Key Project No. 22JC1401500 (T. L.), Shanghai Municipal of Science and Technology Major Project No. 2021SHZDZX0102, and the HPC of School of Mathematical Sciences and the Student Innovation Center, and the Siyuan-1 cluster supported by the Center for High Performance Computing at Shanghai Jiao Tong University.
\section*{Statements and Declarations}
The authors have no conflicts of interest to declare. 

% F-Principle is a type of implicit bias of DNNs, which is widely studied \citep{razin2020implicit}.

% The F-Principle focuses on the response frequency, which also inspires
% the study of the frequency bias in the image frequency. \citet{chakrabarty2019spectral}
% found in the image recovery task, deep image prior \citet{ulyanov2018deep},
% the network recover the low frequency information first. \citet{wang2019high}
% shows that high-frequency component of images helps explain the generalization
% of convolutional neural networks. More detailed empirical and theoretical
% study for the frequency bias among the input variables (image frequency)
% is important.

% =============================

% \citet{dong2019distillation} show that neural network tends to fit
% the infomrative information (i.e. the eigenspaces associated with
% the largest few eigenvalues of NTK) first and the non-informative
% information later. As shown in \citet{zhang2019explicitizing,yang2019fine,cao2019towards},
% each eigen function corresponds to a certain frequency exactly. The
% study of \citet{dong2019distillation} is consistent with the F-Principle.
% Based on this bias study, \citet{dong2019distillation} proposed a
% self-distillation algorithm for noisy label refinery. 

% \bibliographystyle{elsarticle-harv}
\bibliography{DLRef}% common bib file

\end{document}